%% file: main.tex
\documentclass[10pt,journal,compsoc]{IEEEtran}

\ifCLASSOPTIONcompsoc
  \usepackage[caption=false,font=footnotesize,labelfont=sf,textfont=sf]{subfig}
\else
  \usepackage[caption=false,font=footnotesize]{subfig}
\fi

\usepackage{graphicx}
\graphicspath{{imgs/}}                 % images under imgs/
\usepackage{amsmath,amssymb,mathtools}
\usepackage{booktabs,multirow}
\usepackage{siunitx}
\usepackage[caption=false,font=footnotesize]{subfig}  % IEEE suggests subfig
\usepackage{url}
\usepackage{color}
\usepackage{balance}
\usepackage{tabularx}
\usepackage{booktabs}
\usepackage{multirow}
\usepackage{makecell,array}
\usepackage{ragged2e}

\usepackage{bibunits}

\usepackage[font=footnotesize]{caption}
\usepackage{cite} % For compsoc this will be non-compressed by default
\usepackage{xcolor}
\usepackage[colorlinks,linkcolor=blue]{hyperref}

\title{An Anatomy of Vision-Language-Action Models: From Modules to Milestones and Challenges}

\author{
Chao Xu, Suyu Zhang, Yang Liu, Baigui Sun, Weihong Chen, Bo Xu, Qi Liu, Juncheng Wang, \\Shujun Wang, Shan Luo, Jan Peters, Athanasios V. Vasilakos, Stefanos Zafeiriou, Jiankang Deng
\thanks{C. Xu, S. Zhang, Y. Liu, B. Sun, W. Chen, B. Xu, Q. Liu are with IROOTECH TECHNOLOGY (e-mail: chaoxuxc@gmail.com, sunbaigui85@gmail.com).}
\thanks{C. Xu, S. Zhang, Y. Liu, B. Sun are with Wolf 1069 b Lab, Sany Group.}
\thanks{Y. Liu and S. Luo are with the Department of Engineering, King’s College London (e-mail: yang.15.liu@kcl.ac.uk, shan.luo@kcl.ac.uk).}
\thanks{J. Wang and S. Wang are with the Hong Kong Polytechnic University (e-mail: wjc2830@gmail.com, shu-jun.wang@polyu.edu.hk).}
\thanks{J. Peters is with the Computer Science Department of the Technische Universität Darmstadt (e-mail: peters@ias.tu-darmstadt.de).}
\thanks{A. Vasilakos is with
Department of ICT and Center for AI Research, University of Agder (UiA) (e-mail: th.vasilakos@gmail.com).}
\thanks{S. Zafeiriou and J. Deng are with the Department of Computing, Imperial College London (e-mail: j.deng16@imperial.ac.uk, s.zafeiriou@imperial.ac.uk).}
\thanks{C. Xu, S. Zhang and Y. Liu contributed equally to this work.}
}

\IEEEtitleabstractindextext{%
\input{sections/1_abstract} % <-- put \begin{abstract}...\end{abstract} inside this file

}

\begin{document}
\maketitle
\bstctlcite{IEEEexample:BSTcontrol} % 启用上面的控制项
\IEEEdisplaynontitleabstractindextext  % display abstract on first page
\IEEEpeerreviewmaketitle               % safe for review stage (no harm if final)

% ---------- Main Sections ----------
\input{sections/2_introduction}

\input{sections/3_methods}

\input{sections/4_timeline}

\input{sections/5_challenges}
\input{sections/7_conclusion}

% ---------- Acknowledgments (optional) ----------
% \ifCLASSOPTIONcompsoc
%   \section*{Acknowledgments}
% \else
%   \section*{Acknowledgment}
% \fi
% This work was supported by ...

% ---------- Appendix (optional) ----------
% \appendices
% \section{Additional Results}
% \input{sections/7_appendices} % create this file only if needed

% ---------- Bibliography ----------
% \newpage
\bibliographystyle{IEEEtran}
\bibliography{references}

% \begingroup
%   \sloppy
%   \setlength{\emergencystretch}{3em}   % 关键：允许额外拉伸，避免溢出

%   \bibliographystyle{IEEEtran}
%   \bibliography{references}

% \endgroup
% \clearpage
% \begin{bibunit}[IEEEtran]
%     \input{sections/X_supp}
%     \putbib[references]  % 指定用 references.bib
% \end{bibunit}

\input{sections/X_supp}

% ---------- Bio (TPAMI usually after references; remove if not needed) ----------
% \begin{IEEEbiography}[{\includegraphics[width=1in,height=1.25in,clip,keepaspectratio]{author1}}]{First Author}
%   Biography text.
% \end{IEEEbiography}

\balance
\end{document}

%% file: sections/1_abstract.tex
\begin{abstract}
Vision-Language-Action (VLA) models are driving a revolution in robotics, enabling machines to understand instructions and interact with the physical world. This field is exploding with new models and datasets, making it both exciting and challenging to keep pace with. This survey offers a clear and structured guide to the VLA landscape. We design it to follow the natural learning path of a researcher: we start with the basic Modules of any VLA model, trace the history through key Milestones, and then dive deep into the core Challenges that define recent research frontier.
Our main contribution is a detailed breakdown of the five biggest challenges in: (1) Representation, (2) Execution, (3) Generalization, (4) Safety, and (5) Dataset and Evaluation. This structure mirrors the developmental roadmap of a generalist agent: establishing the fundamental perception-action loop, scaling capabilities across diverse embodiments and environments, and finally ensuring trustworthy deployment—all supported by the essential data infrastructure. For each of them, we review existing approaches and highlight future opportunities. We position this paper as both a foundational guide for newcomers and a strategic roadmap for experienced researchers, with the dual aim of accelerating learning and inspiring new ideas in embodied intelligence. A live version of this survey, with continuous updates, is maintained on our \href{https://suyuz1.github.io/VLA-Survey-Anatomy/}{project page}.
\end{abstract}

%% file: sections/2_introduction.tex
\section{Introduction}

\input{imgs/overview}
The quest for general-purpose robots that can operate in real-world human environments is a central goal of artificial intelligence. In recent years, a new approach has emerged as one of the most promising paths toward this goal: Vision-Language-Action (VLA) models. By connecting vision, language, and physical action, these models have catalyzed rapid progress, making the field of embodied intelligence both exciting and increasingly complex.

To help navigate this rapidly growing landscape, numerous survey papers have recently emerged, covering the field from various perspectives. On the one hand, several works provide focused, in-depth reviews on specific technical subareas, such as action tokenization~\cite{zhong2025survey}, efficient training paradigms~\cite{guan2025efficient}, and post-training methodologies~\cite{xiang2025parallels}, offering granular insights into individual system components. On the other hand, broader surveys~\cite{zhang2025pure, din2025vision, sapkota2025vision, shao2025large, kawaharazuka2025vision, ma2024survey} offer comprehensive system overviews. These works typically serve as structured taxonomies, organizing the VLA landscape by model architectures, input modalities, or training objectives, providing readers with a systematic list of the core components.

However, we identify two key gaps that this survey aims to address. First, existing surveys often relegate research challenges to a concluding section—a high-level overview appended at the end of the paper. The field still lacks a unified resource that places these challenges at its core, systematically breaking them down, comparing alternative solution paths, and charting clear directions for future work. For researchers aiming to make novel contributions, a mere list of problems is insufficient; what is needed is a deep, structured analysis of the problem space.
Second, the structure of most surveys does not align how researchers learn a new field. Most existing works simply list and group methods by category—like grouping visual-based approaches in one chapter and control strategies in another. While this facilitates quick reference, it presents a fragmented view of the field. It provides extensive information but fails to illustrate how these pieces integrate into a coherent, evolving research timeline. Consequently, such surveys do not guide newcomers from foundational concepts to recent breakthroughs along a clear, progressive learning trajectory.

This survey makes two core contributions to address these gaps.
Our primary contribution is a deep and systematic analysis of the core challenges in VLA research. Rather than appearing as a brief concluding section, our challenge analysis forms the central pillar of this survey. We identify five key challenges following the developmental roadmap of VLA: (1) Multi-Modal Alignment and Physical World Modeling, (2) Instruction Following, Planning, and Robust Real-Time Execution, (3) From Generalization to Continuous Adaptation, (4) Safety, Interpretability, and Reliable Interaction, (5) Data Construction and Benchmarking Standards. For each, we provide an in-depth review of competing solutions and outline concrete avenues for future research. Our goal is twofold: to help researchers efficiently navigate the vast landscape of existing work and to position this section as a direct catalyst for novel research ideas.

Our second contribution is the unique structure of this survey, designed to mirror the natural learning journey of a researcher. We intentionally structure this survey as a step-by-step roadmap. We begin with a detailed breakdown of the foundational \textit{Modules} that constitute any VLA model, establishing a shared vocabulary. We then trace the historical evolution through key \textit{Milestones}, providing context for how the field has arrived at its current state. This journey culminates in our deep dive into the core \textit{Challenges},  demonstrating recent trends and pointing out future directions. This structure allows newcomers to build expertise from the ground up, while allowing experienced researchers to access the sections most relevant to their interests. The structure of this survey is illustrated in Fig.~\ref{fig:overview}.
This work is designed as a living resource, and \href{https://suyuz1.github.io/Survery/}{project page} will be \textit{continuously} updated to reflect advances at the research frontier.

%% file: imgs/overview.tex
\begin{figure}[htbp]
    \centering
    \includegraphics[width=0.8\columnwidth]{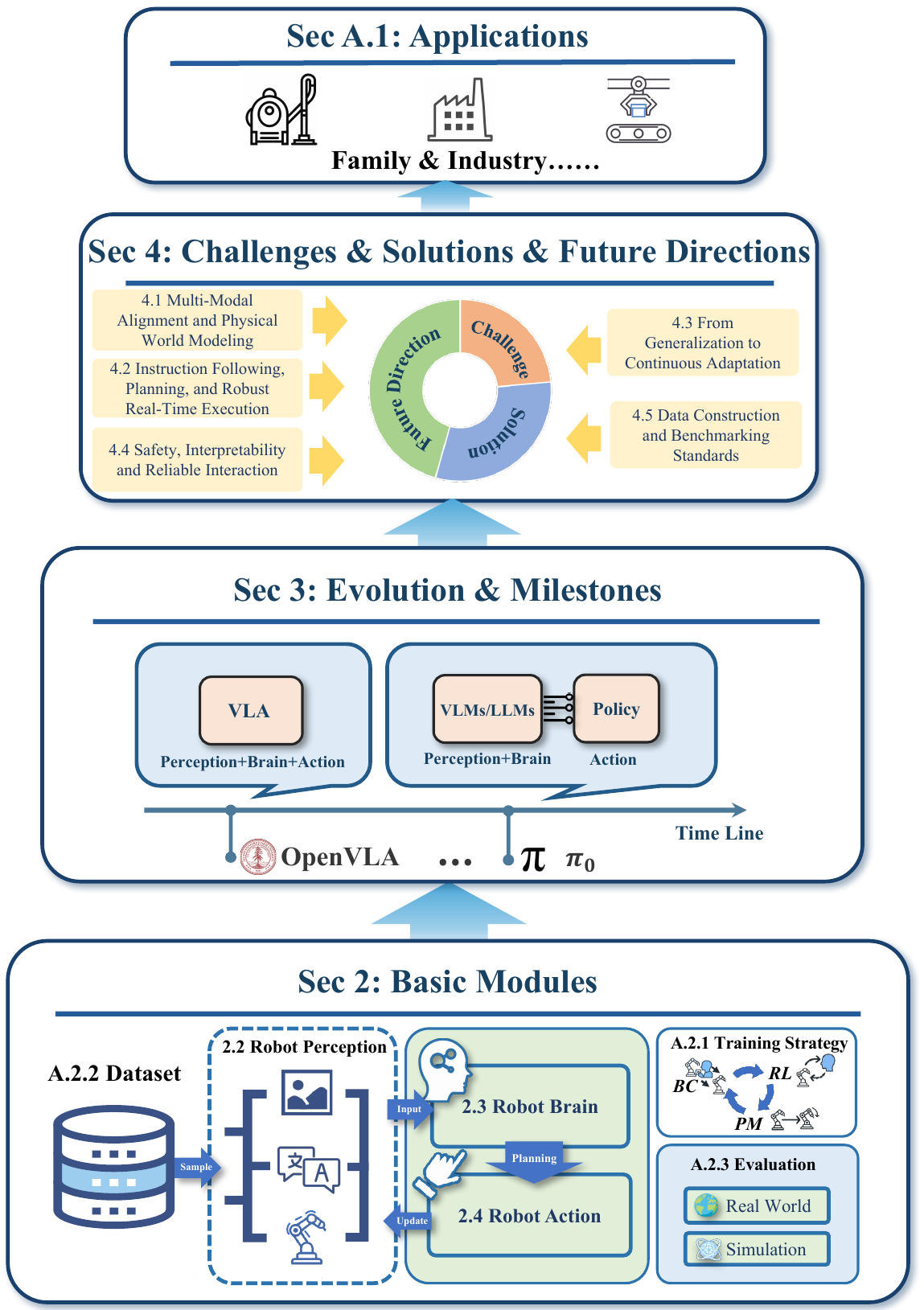}
    \caption{\textbf{The structure of this survey in a pyramid format}. Section~\ref{sec:2} lays the foundational knowledge by deconstructing the core components of any VLA model. Building upon this, the second stage, Section~\ref{sec:3}, traces the historical evolution of the field through its most representative works, providing context and intuition. The deepest stage, Section~\ref{sec:4}, serves as the intellectual core, offering an in-depth analysis of the grand open problems and outlining actionable future research directions. The final section depicts the various applications, which are included in Appendix~\textcolor{blue}{A.1}. 
    }
    \label{fig:overview}
\end{figure}

%% file: sections/3_methods.tex
\section{Basic Modules}
\label{sec:2}

\subsection{Overall and Architectural Trend}
\label{sec:2.1}
Vision-Language-Action (VLA) systems integrate perception, reasoning, and control to translate abstract instructions into physical actions. Typically, a VLA system comprises three core modules: the perception module extracts grounded observations, the brain module fuses multimodal inputs for planning, and the action module executes motor commands. 
Recently, these components are undergoing a fundamental shift: Perception (Sec.~\ref{sec:2.2}) is evolving from standard visual backbones to Language-Aligned Transformers (e.g., SigLIP) to bridge the semantic gap, increasingly augmented by geometric representations (e.g., DINOv2) to ensure manipulation precision. The Brain (Sec.~\ref{sec:2.3}) is converging toward pre-trained VLMs, leveraging internet-scale knowledge to enable zero-shot generalization and unified token processing. Finally, Action (Sec.~\ref{sec:2.4}) is pivoting from discrete tokenization towards continuous generative modeling (e.g., Diffusion), achieving smooth, multi-modal distribution modeling. Notably, to prioritize the in-depth analysis of challenges (Sec.~\ref{sec:4}), we provide a streamlined overview here due to limited space. For detailed architectural taxonomies, we recommend other specialized surveys~\cite{kawaharazuka2025vision, zhang2025pure}.

\subsection{Robot Perception}
\label{sec:2.2}
\subsubsection{Vision Encoders in VLA}
\textbf{(1) Convolutional Networks (CNNs).} 
CNNs~\cite{cnn} remain indispensable in VLA due to their strong local feature extraction and translation equivariance, making them effective visual encoders in real-time and resource-constrained settings. Modern architectures such as ResNet and EfficientNet~\cite{resnet,efficientnet} are widely adopted. CNNs commonly serve as visual backbones in end-to-end policies by encoding RGB or depth observations into compact features for downstream decision-making; representative systems such as Diffusion Policy~\cite{diffusion_policy} and SPECI~\cite{speci} use ResNet-based encoders. 
CNNs also integrate naturally into hierarchical designs, where lightweight models handle high-frequency perception, as in HiRT~\cite {hirt,efficientnet}, which employs EfficientNet-B3. As world-model-based VLA frameworks grow in complexity, CNNs increasingly act as compact encoders for high-dimensional observations. For example, LUMOS~\cite{lumos} uses a CNN front-end to produce latent features consumed by RSSM~\cite{RSSM} for prediction and planning.

\noindent\textbf{(2) Vision Transformers (ViT).} 
ViT~\cite{vit} and its variants have become the dominant perception backbone in modern VLA systems. Their self-attention captures global context and long-range dependencies, and patch tokenization aligns visual inputs with Transformer-based language models, making ViT well suited for end-to-end VLA pipelines~\cite{cnn_vs_vit}. Contemporary VLA frameworks therefore rely heavily on large-scale pretrained ViT encoders, typically fine-tuned for stronger generalization and efficiency. ViT-based visual encoders in VLA generally follow four structural paradigms:

\textbf{a) Language-Supervised Visual Encoders.}
Models such as CLIP~\cite{clip} and SigLIP~\cite{siglip} learn vision features aligned to human semantics via contrastive learning from internet-scale image–text pairs. 
Adopting such encoders is now standard practice: for example, $\pi_{0}$~\cite{pi0}, RDT-1B~\cite{rdt1b}, TriVLA~\cite{trivla}, and ForceVLA~\cite{forcevla} use SigLIP as their vision backbone, while many others rely on CLIP, e.g., DeeR-VLA~\cite{deer_vla}, RationalVLA~\cite{rationalvla}, MinD~\cite{mind}. 
Some works innovate in their use, for instance, OTTER~\cite{otter} extracts features from the final layer of a frozen CLIP ViT to obtain strongly language-aligned visual representations.

\textbf{b) Self-Supervised Visual Encoders.}
These models, exemplified by DINOv2~\cite{dinov2}, avoid textual labels and learn robust visual representations from large unlabeled corpora, capturing fine-grained geometry and spatial structure that make them particularly effective for contact-rich manipulation tasks requiring precise physical cues.
For example, LexVLA~\cite{lexvla} employs a frozen DINOv2 encoder and a lightweight adapter to map local visual features into sparse, language-aligned lexical representations.

\textbf{c) Hybrid Architectures.} To combine the semantic strengths of language-supervised encoders with the geometric precision of self-supervised ones, an increasingly common strategy is to adopt a hybrid approach. 
Recent VLA frameworks, including OpenVLA~\cite{openvla}, OpenVLA-OFT~\cite{openvla_oft}, GraspVLA~\cite{graspvla}, UniVLA~\cite{univla}, and VLA-RL~\cite{vla_rl} often employ a SigLIP$+$DINOv2 hybrid to attain strong performance on both semantic understanding and geometric reasoning.

\textbf{d) Vision-Language Models (VLMs).} The most integrated paradigm directly adopts pretrained VLMs as high-level visual encoders, producing language-conditioned visual embeddings rather than raw pixel features.
Examples include PaLI-X~\cite{pali_x} in RT-H~\cite{rt_h}; PaliGemma~\cite{paligemma} in Hume~\cite{hume} and Hi Robot~\cite{hirobo}; Qwen-VL~\cite{qwen_vl} in VTLA~\cite{vtla}, CombatVLA~\cite{combatvla}, and OpenHelix~\cite{openhelix}, which leverage VLMs’ fused vision–language context to provide higher-level inputs for policy and planning.

\subsubsection{Language Encoders in VLA}

Language instructions form the semantic core of VLA systems, defining task objectives and providing high-level context.  The language encoder has evolved alongside advances in natural language processing, and in practice falls into three main categories:

\noindent\textbf{(1) Transformer-Based Language Encoders.} The earliest approach involves the use of standard Transformer-based language encoders. These VLA systems adopt text-only Transformers (e.g., BERT~\cite{bert}, T5~\cite{t5}) pretrained on large corpora to encode instructions, providing strong semantics as the entry point of the control stack. 
Classical examples include RDT-1B~\cite{rdt1b} with T5-XXL~\cite{t5}, RoboBERT~\cite{robobert} with BERT, and early partial implementations of Octo~\cite{octo} that rely on such modules.

\noindent\textbf{(2) Large Language Models (LLMs).} With the rise of LLMs, VLA systems increasingly adopt billion-parameter models as their language backbone, leveraging their richer world knowledge and commonsense reasoning to interpret ambiguous and compositional instructions.
Representative choices include Llama-family models (e.g., OpenVLA-OFT~\cite{openvla_oft}, VLA-RL~\cite{vla_rl} with Llama-2~\cite{llama2} 7B; HiRT~\cite{hirt} with Llama-based InstructBLIP~\cite{instructblip}), Gemma-family models~\cite{gemma} (e.g., $\pi_0$~\cite{pi0}/$\pi_{0.5}$~\cite{pi0_5} with Gemma~2B~\cite{gemma2}), and InternLM2~\cite{internlm2} (e.g., GraspVLA~\cite{graspvla}).

\noindent\textbf{(3) Vision-Language Models (VLMs).} 
The recent trend is to adopt native VLMs, where the language module is no longer a standalone component but is jointly pretrained with vision for end-to-end multimodal understanding. 
For example, several VLA systems explicitly adopt well-known vision–language models:
DeeR-VLA~\cite{deer_vla} and RoboFlamingo~\cite{roboflamingo} build on OpenFlamingo~\cite{openflamingo};
Diffusion-VLA~\cite{diffusion_vla} instead  employs Qwen-VL~\cite{qwen_vl} while
MemoryVLA~\cite{memoryvla} is developed upon the 7B Prismatic VLM~\cite{prismatic_vlm}.
In contrast, Dexbotic~\cite{dexbotic} pretrains its own dedicated model, DexboticVLM, tailored to dexterous manipulation.
Other systems including  InstructVLA~\cite{instruct_vla}, FlowVLA~\cite{flowvla}, and others, also utilize VLMs as their language encoders.

\subsubsection{Proprioceptive Encoders in VLA}
Proprioceptive inputs are provided by onboard sensors and typically include 
(i) joint states: per-joint position, velocity, and effort/torque; 
(ii) end-effector states: the 6-DoF pose $(x,y,z,\mathrm{roll},\mathrm{pitch},\mathrm{yaw})$, optionally with linear/angular velocities, in the world or base frames~\cite{Lim_2023,ta_vla}; 
and (iii) gripper status: opening width/state and applied force. These data are low-dimensional, structured vectors.

Given the low-dimensional, structured nature of proprioception, MLPs are the standard, efficient encoders, whose outputs are fused with vision and language via concatenation or conditioning (e.g., FiLM~\cite{film}). Many VLA models follow this design. TriVLA~\cite{trivla} employs an embodiment-specific MLP, RDT-1B~\cite{rdt1b} encodes low-D robot states with an MLP, SPECI~\cite{speci} trains an MLP from scratch on joint angles and gripper states, and systems such as OpenVLA-OFT~\cite{openvla_oft} and the GR series~\cite{gr_1,gr_2} similarly include MLP modules for proprioceptive fusion.

\subsection{Robot Brain}
\label{sec:2.3}
The robot brain is the core of a VLA system, responsible for fusing multimodal representations from input modules, performing reasoning and planning, and ultimately generating action intentions. Current architectures primarily follow four mainstream technical directions:

\noindent\textbf{(1) Transformer.}
The Transformer serves as a core VLA architecture by tokenizing vision, language, and proprioception inputs and using self-attention to fuse multimodal tokens and learn an end-to-end perception-to-action mapping.
A Generalist Agent~\cite{generalist_agent} demonstrates the capacity of a decoder-only Transformer to handle multiple modalities and tasks, and models such as VIMA~\cite{vima} and GR-1/GR-2~\cite{gr_1,gr_2} further adopt Transformer-based generalist policies. Other approaches, such as SPECI~\cite{speci}, apply temporal Transformers across both high-level reasoning and low-level execution. Beyond Transformers, recent alternatives also emerge; 
RoboMamba~\cite{robomamba} adapts the Mamba~\cite{mamba} architecture to VLA for more efficient long-sequence processing.

\noindent\textbf{(2) Diffusion Transformer (DiT).}
Unlike Transformer-only policies that predict actions directly, this paradigm uses a diffusion model as the generative core, with a Transformer guiding the denoising process. Diffusion models are well suited for robot control because they model complex continuous distributions and produce smooth, natural motion trajectories. 
Diffusion Policy~\cite{diffusion_policy} provides early evidence of the effectiveness of denoising-based generation, helping establish diffusion as a strong policy-learning paradigm. More recent methods, such as RDT-1B~\cite{rdt1b} and TriVLA~\cite{trivla}, integrate diffusion on top of Transformer backbones to map semantics to actions through multi-step denoising.

\noindent\textbf{(3) Hybrid Architectures.}
These models pair Transformer-based semantic reasoning with a diffusion~\cite{diffusion} or flow-matching~\cite{flowmatching} head for high-frequency, smooth control. $\pi_0$~\cite{pi0} exemplifies this design by using a pretrained VLM as the Transformer backbone for perception and a separate Flow Matching head for action generation. Octo~\cite{octo} and ConRFT~\cite{conrft} follow a similar pattern, combining a Transformer backbone with a generative action head. 
Diffusion-VLA~\cite{diffusion_vla} injects LLM reasoning into the diffusion process to coordinate high-level planning with low-level execution.
MinD~\cite{mind} adopts a hierarchical hybrid structure, using distinct diffusion models for low-frequency video prediction and high-frequency action control.

\noindent\textbf{(4) Vision-Language Models (VLMs).} 
This paradigm treats a full pretrained vision-language model as the core robot brain, leveraging its perception, multimodal fusion, commonsense reasoning, and sequence modeling, while integrating robot-specific proprioception and action spaces on top.
RT-2~\cite{rt_2} is a milestone in this direction, extending the VLM’s (i.e., PaLI-X~\cite{pali_x}/PaLM-E~\cite{palm_e}) output space to include action tokens, effectively creating an embodied agent. Nearly all current SOTA VLA models, including OpenVLA~\cite{openvla}, $\pi_{0.5}$~\cite{pi0_5}, CoT-VLA~\cite{cot_vla}, SafeVLA~\cite{safevla}, DeeR-VLA~\cite{deer_vla}, GraspVLA~\cite{graspvla}, VTLA~\cite{vtla}, UniVLA~\cite{univla}, VLA-RL~\cite{vla_rl}, WorldVLA~\cite{worldvla}, TraceVLA~\cite{trace_vla}, PointVLA~\cite{pointvla}, 3D-VLA~\cite{3d_vla}, and BridgeVLA~\cite{bridgevla} build their decision-making on strong pretrained VLMs.
In hierarchical systems, VLMs often act as high-level planners or span both high- and low-level policies, as in A Dual Process VLA~\cite{dual_process_vla}, Hi Robot~\cite{hirobo}, HAMSTER~\cite{hamster}, and HiRT~\cite{hirt}.

\subsection{Robot Action}
\label{sec:2.4}
Robot action is the VLA system’s final execution interface, translating abstract decisions from the robot brain into concrete, low-level control commands. Its design directly determines action precision, smoothness, real-time performance, and generalization.

\subsubsection{Action Representation}
Action space representation defines the target language that the model predicts.
Representing typically high-dimensional, continuous robot actions involves a key trade-off between performance and learnability.

\noindent\textbf{(1) Discrete Spaces.}
Continuous controls are discretized into bins and cast as a next-token classification problem, naturally reusing Transformer stacks for sequence prediction. 
This is common in generalist Transformer agents (e.g., A Generalist Agent~\cite{generalist_agent}, VIMA~\cite{vima}, RT-H~\cite{rt_h}, SafeVLA~\cite{safevla}) and in many recent VLA systems (e.g., UniVLA~\cite{univla}, VLA-RL~\cite{vla_rl}, WorldVLA~\cite{worldvla}, TraceVLA~\cite{trace_vla}, CombatVLA~\cite{combatvla}).

\noindent\textbf{(2) Continuous Spaces.}
Actions are regressed directly in normalized continuous domains (e.g., joint angles, end-effector velocities), yielding smoother, higher-precision control at the cost of high demands on model learning ability. 
This aligns naturally with diffusion or flow-matching policies (e.g., Diffusion Policy~\cite{diffusion_policy}, TriVLA~\cite{trivla}, RDT-1B~\cite{rdt1b}, $\pi_{0}$~\cite{pi0}) and with continuous variants of prior discrete models (e.g., OpenVLA-OFT~\cite{openvla_oft}). Other systems such as iRe-VLA~\cite{ire_vla}, GraspVLA~\cite{graspvla}, and Hume~\cite{hume} also adopt continuous control.

\noindent\textbf{(3) Hybrid Spaces.}
To combine strengths, hybrids assign discrete and continuous encodings to different control facets: BridgeVLA~\cite{bridgevla} uses continuous translation with discretized rotation. HiRT~\cite{hirt} treats EE pose as continuous while gripper open/close is discrete. Hierarchical models often keep high-level skills discrete and low-level execution continuous (e.g., Hi Robot~\cite{hirobo}, HAMSTER~\cite{hamster}, $\pi_{0.5}$~\cite{pi0_5}).

\subsubsection{Action Decoding}

\textbf{(1) Autoregressive Decoding.}
In autoregressive (AR) decoding, the policy emits actions step by step with causal masking, and each prediction conditions on all previously generated actions and observations, enabling modeling of long-range temporal dependencies.
AR remains standard in early and many recent VLA models (e.g., A Generalist Agent~\cite{generalist_agent}, VIMA~\cite{vima}, RT-H~\cite{rt_h}, SafeVLA~\cite{safevla}, GR-2~\cite{gr_2}, 3D-VLA~\cite{3d_vla}, UniVLA~\cite{univla}, OpenVLA~\cite{openvla}, TraceVLA~\cite{trace_vla}, CombatVLA~\cite{combatvla}).

\noindent\textbf{(2) Non-Autoregressive Decoding.}
To reduce latency, non-AR decoders predict an action horizon in one or a few passes. 
One path replaces causal attention with bidirectional attention to infer all steps jointly (e.g., OpenVLA-OFT~\cite{openvla_oft}). 
Another uses inherently non-AR generators such as diffusion or flow matching that iteratively denoise or transform the whole sequence in parallel (e.g., Diffusion Policy~\cite{diffusion_policy}, TriVLA~\cite{trivla}, RDT-1B~\cite{rdt1b}, $\pi_{0}$~\cite{pi0}, RoboBERT~\cite{robobert}, Hume~\cite{hume}, DeeR-VLA~\cite{deer_vla}).

\noindent\textbf{(3) Hybrid Decoding.}
A practical compromise is chunking: the policy operates autoregressively over coarse time (emitting chunks), but non-autoregressively within each chunk (parallel refinement), which improves both stability and throughput. 
A representative example is $\pi_{0.5}$~\cite{pi0_5}, which performs AR semantic decisions with parallel low-level chunk generation. CoT-VLA~\cite{cot_vla}, UniVLA~\cite{univla}, and WorldVLA~\cite{worldvla} follow the same design, which support long-horizon coherence with efficient local rollout.

%% file: sections/4_timeline.tex
\section{Evolution \& Milestones}
\label{sec:3}
\input{imgs/timeline}
The evolution of Vision-Language-Action (VLA) models is driven by the need to overcome the brittleness of traditional modular pipelines and achieve the broad generalization seen in foundation models.
This evolution reflects a steady shift from passive multimodal perception to active, embodied reasoning and control. An overview of VLA milestones is shown in Fig.~\ref{fig:timeline} and Appendix Tab.~\textcolor{blue}{S3}.

From 2017 to 2019, the Vision-and-Language Navigation (VLN) benchmark~\cite{vln} pioneers large-scale evaluation of agents aligning linguistic instructions with visual environments for physical navigation. EmbodiedQA~\cite{embodiedqa} advances this direction by defining embodied intelligence through a closed perception–action loop, establishing an early theoretical foundation. Follow-up work such as BabyAI~\cite{babyai}, RCM~\cite{rcm}, and Point-Cloud EQA~\cite{point_cloud_eqa} further refine the paradigm by improving language-to-action learning and introducing early forms of 3D geometric reasoning.

The period from 2020 to 2021 marks a shift toward \emph{long-horizon reasoning and language-conditioned embodied control}. ALFRED~\cite{alfred} introduces the first interactive benchmark combining high-level goals, step-by-step instructions, and object–environment interactions, establishing realistic long-horizon tasks. ALFWorld~\cite{alfredworld} extends this direction by linking symbolic reasoning with visually grounded execution, and BEHAVIOR~\cite{behavior} standardizes long-horizon household evaluation in high-fidelity simulation. A pivotal milestone of this era is CLIPort~\cite{cliport}, which integrates pretrained visual representations into a language-conditioned policy, demonstrating that internet-scale knowledge enables zero-shot generalization in robotic manipulation.

Since 2022, VLA enters the era of \emph{large models and generalized learning}.  
SayCan~\cite{saycan} is the first to introduce a hierarchical framework that separates LLM-based high-level planning from low-level skill execution, using affordance and value estimates from the robot to ground candidate subtasks and select feasible actions. 
Inner Monologue~\cite{inner_monologue} for the first time embeds language models within continuous multimodal feedback loops, achieving self-reflection and dynamic behavioral adjustment.
RT-1 and RT-2 ~\cite{rt_1,rt_2} realize end-to-end learning from vision and language to action via Transformer architectures, marking the birth of a truly unified VLA framework.

In 2023, multiple advances emerge, most notably in \emph{unified multimodal backbones, generative action modeling, and cross-embodiment data scaling}.
PaLM-E~\cite{palm_e} embeds visual and state representations directly into pretrained LLMs, achieving for the first time a unified multimodal input space.
The introduction of Diffusion Policy~\cite{diffusion_policy} applies generative diffusion models to action modeling, bringing greater stability and expressiveness to high-dimensional continuous control, and marking a key paradigm shift in policy generation for VLA.
Open X-Embodiment~\cite{oxe} represents a meaningful milestone in robotic learning, providing large-scale and diverse cross-robot data with open access, and driving the field toward more general and powerful embodied models.

Building on the previous year's breakthroughs, 2024 broadens the frontier across \emph{open-source scaling, generalist policies, flow/denoising action generation, web-scale video pretraining, and 3D world modeling}.
Octo~\cite{octo} establishes a generalist policy capable of cross-platform, multi-task control.
OpenVLA~\cite{openvla} becomes the first fully open-source 7B VLA model, lowering the barrier for large-scale research and deployment.
$\pi_0$~\cite{pi0} is the first to combine pretrained VLMs with flow-matching action generation, setting a new architectural reference point for general and precise control.
GR-2~\cite{gr_2} systematizes web-scale generative video pretraining for VLA, enabling broad generalization without proportional robot labels.
3D-VLA~\cite{3d_vla} marks a shift toward full 3D world modeling by coupling a generative 3D world model with VLA for plan-by-imagination.

By 2025, VLA research enters a stage of pluralistic evolution, where diverse embodiments, modalities, and learning paradigms co-evolve toward \emph{general robotic intelligence}.
Humanoid-VLA~\cite{humanoid_vla} and GR00T N1~\cite{groot_n1} extend VLA to full-body humanoid control. Another direction targets open-world autonomy, emphasizing deeper understanding and reasoning.
PointVLA~\cite{pointvla} injects point-cloud features without retraining the core model, enabling faithful 3D understanding for open-world settings.
Cosmos-Reason1~\cite{cosmos_r1} is the first to standardize physically grounded reasoning for VLAs, unifying ontologies and benchmarks into an open reasoning pipeline and shifting the field toward plug-and-play, physics-constrained reasoning.
CoT-VLA~\cite{cot_vla} introduces the first explicit visual chain-of-thought, predicting subgoal images as intermediate reasoning before action generation.
At the core, some models aim to unify prior advances by integrating hierarchy, reasoning, and control.
$\pi_{0.5}$~\cite{pi0_5} unifies high-level reasoning and low-level control via hierarchical Transformers, enabling long-horizon operation without target-specific robot data.
LUMOS~\cite{lumos} integrates a learned world model with on-policy RL into a single system.
VLA-RL~\cite{vla_rl} scales online RL to pretrained VLAs, addressing imitation learning’s OOD limitations.
GEN-0~\cite{gen0} offers early evidence for scaling laws in robotics, showing that large-scale interaction data enables phase transitions in cross-embodiment generalization.

%% file: imgs/timeline.tex
\begin{figure*}[t]
    \centering
    \includegraphics[width=0.95\textwidth]{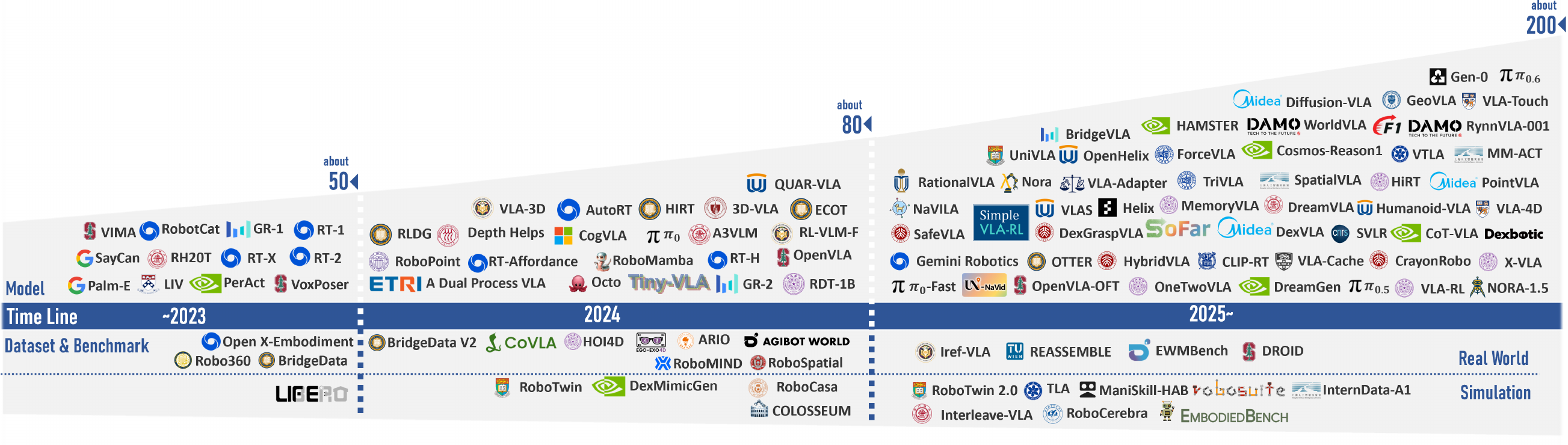}
    \caption{\textbf{The timeline of VLA models, datasets, and evaluation benchmarks from 2022 to 2025}. The top row presents major VLA models introduced each year. The bottom row displays key datasets used to train and benchmarks to evaluate these models, grouped by release year.}
    \label{fig:timeline}
\end{figure*}

%% file: sections/5_challenges.tex
\section{Challenges \& Solutions \& Future Directions}
\label{sec:4}
Fig.~\ref{fig:Challenges_overview} provides an overview of the five core challenges addressed in this section, along with their respective sub-challenges and the relevant papers involved.

\input{imgs/challenges_overall}
\subsection{Multi-Modal Alignment and Physical World Modeling}
\label{sec:4.1}
Fig.~\ref{fig:challenge_1} illustrates the three levels of this challenge, which are elaborated in detail below.
 
\subsubsection{The GAP between Semantics, Perception, and Physical Interaction}
\label{sec:4.1.1}

\input{imgs/challenge_1}
Vision-Language-Action (VLA) tasks center on three core components: vision for perceiving the world, language for conveying high-level instructions, and action for interacting with the physical environment.  Together, they form an integrated embodied framework linking perception, reasoning, and execution.
The central challenge is bridging the gap between abstract semantics and grounded physical reality, which can be decomposed into three subproblems: 

\noindent\textbf{(1) Vision Language Gap.} 
Vision provides high-dimensional perceptual input, while language offers abstract symbolic semantics. Establishing a precise mapping between these distinct modalities is essential for grounding visual understanding and goal reasoning in the physical world~\cite{openhelix,semantic_vla}. 
Some approaches address this challenge by \emph{enhancing visual representations} to make them more responsive to language conditioning. OTTER~\cite{otter} introduces text-aware feature extraction that preserves semantics aligned with task descriptions, while LIV~\cite{LIV} employs a contrastive framework on robot-control data to construct a joint vision–language embedding space, enabling visual features to become inherently sensitive to linguistic cues. A recent paradigm bridges the vision–language gap via \emph{symbolic reasoning} with natural language as an intermediate representation, powered by LLMs.
ACT-LLM~\cite{actllm} translates visual observations into structured state descriptions for symbolic reasoning.
Look Leap~\cite{look_leap} pushes this further by generating full structured action plans, elevating vision–language alignment to a higher cognitive level and reframing the problem as one of reasoning.

\noindent\textbf{(2) Vision--Language Action Gap.} 
Although multimodal models achieve strong perception–semantics alignment, a gap remains when grounding this understanding into physical action~\cite{knowledge_insulating_vla}.
One direction is \emph{end-to-end fine-tuning}, which reformulates control as sequence generation by discretizing the action space into tokens and fine-tuning a VLM to generate these action tokens in the same way it generates words. RT-2~\cite{rt_2} demonstrates the feasibility of this approach, and subsequent works such as Prompt-a-Robot-to-Walk~\cite{promptrobotwalk}, Grounding MLLMs in Actions~\cite{grounding_mllm}, and OpenVLA~\cite{openvla} adopt similar paradigms.
Another line of work introduces \emph{shared intermediate representations} between language and action. CLIP-RT~\cite{clip_rt} extends vision–language alignment to action generation, and Humanoid-VLA~\cite{humanoid_vla} performs language–action pretraining to narrow the semantic–motor gap. VoxPoser~\cite{voxposer} leverages LLM reasoning to produce intermediate programs and 3D affordance maps that ground perceptual semantics into spatial actions. 
Recent studies further mitigate this mismatch by introducing \emph{hierarchical architectures}~\cite{gemini_robotics, orion, knowledge_insulating_vla} that insert an explicit intermediate layer between language and action, where a VLM serves as a high-level planner and a separate low-level controller executes high-frequency motion.

\noindent\textbf{(3) Multi-modal Sensory Fusion.} 
As VLA systems evolve, perception extends beyond RGB images and language. For precise manipulation, vision and instruction alone are insufficient for accurate physical interaction and fine-grained control~\cite{tla}. Incorporating additional modalities such as tactile, force, and audio sensing is therefore essential for achieving more reliable and comprehensive perception~\cite{beyond_sight, rh20t, touch_vla}, yet it significantly increases the complexity of modality alignment and model optimization.

A common solution is to build \emph{specialized encoders} for each sensory modality and align them with language using contrastive learning. TLA~\cite{tla} integrates tactile perception to improve contact-rich manipulation, and OmniVTLA~\cite{omnivtla} constructs a semantically aligned tactile encoder that links tactile feedback with linguistic concepts. After obtaining effective representations, the challenge shifts to \emph{fusion}, ranging from deep fusion across the full pipeline, as in Tactile-VLA~\cite{tactile_vla}, to modular mixture-of-experts fusion that preserves VLM representations, as in ForceVLA~\cite{forcevla}.
Due to the high cost of collecting real multimodal data, \emph{simulation-based generation} is emerging as a promising alternative. MultiGen~\cite{multigen} explores this direction by generating visual scenes in the simulator and synthesizing additional modalities such as audio to pretrain or enhance real-world policies.

\subsubsection{From 2D Images to Spatial-Temporal Representations}
\label{sec:4.1.2}
Bridging the semantic–perceptual–physical gap requires spatial grounding, meaning that VLA models must accurately capture the 3D structure of the environment. Yet most pretrained VLMs are trained on 2D internet images, creating a core limitation: their reliance on RGB inputs restricts the spatial reasoning needed for real-world robotic operation~\cite{pointvla}. 
Enabling a 2D-native model to acquire spatio-temporal understanding is therefore a central challenge.

\noindent\textbf{(1) Constructing Spatio-Temporal Representations.}
Building spatio-temporal understanding begins with selecting a representation capable of expressing geometric structure and dynamics. Existing approaches primarily follow three directions.
A straightforward option is to augment RGB inputs with \emph{2.5D depth maps}, which provide per-pixel distance information and align naturally with 2D images. Depth Helps~\cite{depth_helps} uses depth as a supervision to learn spatial perception without real sensors, while RoboFlamingo-Plus~\cite{roboflamingoplus} fuses preprocessed depth with RGB features to strengthen spatial awareness. These results show that even simple 2.5D cues can significantly enhance geometric reasoning.
Then, \emph{point clouds} preserve full 3D geometry and offer lossless 3D representation~\cite{geovla}. PointVLA~\cite{pointvla} integrates point cloud inputs into pretrained VLA models to improve spatial reasoning without modifying the backbone. Later systems, such as An Embodied Generalist Agent in a 3D World~\cite{LEO} and GeoVLA~\cite{geovla}, unify 2D and 3D modalities, while FP3~\cite{fp3} rebuilds the perception–decision pipeline around point cloud representations under a pretraining–finetuning paradigm. 
Beyond pure geometry, other studies aim to infuse semantics into point clouds.  SoFar~\cite{sofar} constructs semantic 3D scene graphs, Weakly-Supervised 3D Visual Grounding~\cite{weaklysupervised3dvisualgrounding} transfers 2D–text alignment to 3D by leveraging CLIP, and LMM-3DP~\cite{llm_3dp} fuses back-projected 2D semantic features with geometric point clouds to form unified semantic–geometric representations.
To address the irregular structure of point clouds, other work discretizes 3D space into \emph{voxels} or \emph{occupancy grids}, enabling structured spatial reasoning. OccLLaMA~\cite{occllama} assigns semantic labels to 3D voxels, while RoboMM~\cite{robomm} incorporates multi-view temporal cues to construct unified 3D occupancy grids.
Finally, since real-world operation is dynamic, a static 3D snapshot is insufficient. 
ARM4R~\cite{ARM4R} captures spatio-temporal evolution by predicting the \emph{4D trajectory} of 3D point motion, extending static perception to a time-aware formulation~\cite{trace_vla}.

\noindent\textbf{(2) Architectural Integration.}
Once a spatio-temporal representation is chosen, the next challenge is incorporating geometric information into VLA models without disrupting pretrained alignment.
A common strategy is augmentation and injection through \emph{specialized adapters} that introduce 3D features while preserving the backbone’s integrity as much as possible. PointVLA~\cite{pointvla} directly augments 2D models with point cloud inputs, while GeoVLA~\cite{geovla} processes 2D and 3D streams in parallel. SpatialVLA~\cite{spatialvla} projects 2D semantic features into 3D coordinates using positional encoding and spatial grids to form explicit space–action graphs. 
In contrast, implicit approaches avoid modifying the backbone by attaching external geometric modules, as in Evo-0~\cite{evo0} with VGGT~\cite{vggt}, or by using diffusion-based conditioning to model depth reliability, as in AC-DiT~\cite{ac-dit}.
Another line of work circumvents direct 3D modeling by \emph{reprojecting 3D data into the 2D domain}.
BridgeVLA~\cite{bridgevla} renders point clouds into multi-view images, and OG-VLA~\cite{og_vla} generates orthographic projections to recover 3D pose. Some systems predict in 2D and then lift results into 3D, such as A$^{0}$~\cite{og_vla}, which first predicts interaction points and trajectories in 2D and then lifts them into 3D via depth projection, and RoboPoint~\cite{robopoint}, which back-projects 2D keypoints into 3D to create structured action cues. These methods preserve the strengths of large-scale 2D pretraining while retaining essential 3D awareness.
A third direction avoids explicit reconstruction altogether by relying on the \emph{reasoning ability of large multimodal models}. 
VoxPoser~\cite{voxposer} generates dense voxel-value maps from language-guided code to directly impose linguistic constraints on spatial geometry, and Gemini Robotics~\cite{gemini_robotics} infers 3D structure through large-scale multimodal reasoning. 
Finally, to operationalize the 4D perspective within VLA, recent work \emph{injects tracked motion as temporal context}. 
TraceVLA~\cite{trace_vla} overlays tracked keypoint trajectories as spatial memory, and Spatial Traces~\cite{spatial_traces} fuses tracked points with depth maps to encode structure and motion within a unified input.

\subsubsection{Dynamic and Predictive World Models}
\label{sec:4.1.3}
A truly embodied world representation cannot stop at static geometry or semantics, it must capture dynamics and causality, i.e., construct an internal, predictive world model capable of answering the fundamental question: if the agent executes an action, what happens next?  
Predictive world modeling forms the foundation for counterfactual reasoning, long-horizon planning, and physical understanding.

\noindent\textbf{(1) Representation Space.}
A key design choice is how future states should be represented. 
One option is to predict directly in the observation space by generating future \emph{pixel-level frames}, which provides a high-fidelity, human-interpretable imagination of future states. TriVLA~\cite{trivla} extends video diffusion models for multi-step visual forecasting, while UP-VLA~\cite{up_vla} and CoT-VLA~\cite{cot_vla} generate key subgoal images that indicate the next salient task state. DreamVLA~\cite{dreamvla} enriches prediction with task-critical cues such as dynamic regions, depth, and affordances, and FlowVLA~\cite{flowvla} introduces a visual chain-of-thought mechanism to synthesize physically consistent future scenes. WorldVLA~\cite{worldvla} further models object motion, contact, and state transitions to simulate low-level physical evolution via learned world dynamics.
A complementary approach is prediction in a \emph{latent space}. This strategy first encodes high-dimensional visual observations into a compact, low-dimensional latent space and then learns a simpler model to predict the evolution of this latent state~\cite{v_jepa2}. This is more computationally efficient and robust to irrelevant visual noise. For instance, VLM-in-the-Loop~\cite{vlm_in_loop} explicitly leverages a pretrained latent world model to predict future latent states, while MinD~\cite{mind} proposes a hierarchical world model that performs predictions in dynamic feature spaces at multiple levels of abstraction. WMPO~\cite{wmpo} generates internally in latent space while aligning policy and optimization in pixel space.

\noindent\textbf{(2) Utilization Paradigms.} One paradigm is \emph{policy enhancement}, where the world model is tightly integrated with the policy. Short-term future predictions serve as auxiliary inputs or auxiliary training signals~\cite{f1}, providing the policy with forward-looking intuition for more informed action selection. Most observation-space models, including TriVLA~\cite{trivla}, CoT-VLA~\cite{cot_vla}, and DreamVLA~\cite{dreamvla}, follow this strategy by conditioning their action decoders on predicted future states.
The second paradigm is \emph{explicit planning}, in which the world model serves as a decoupled internal simulator. In this think-before-you-act framework, the agent performs multi-step rollouts of candidate action sequences within the model, evaluates their long-horizon outcomes, and chooses the best plan. This deliberative approach, used in systems such as LUMOS~\cite{lumos}, VLM-in-the-Loop~\cite{vlm_in_loop}, and MinD~\cite{mind}, is particularly effective for tasks requiring long-term foresight and trade-offs.

\subsubsection{Future Directions}
\noindent\textbf{Summary \& Trends}: 
Current VLA architectures struggle with two fundamental disconnects, which existing methods address via a patchwork strategy. (1) Regarding the Modality Disconnect, the prevailing trend relies on modular Late Fusion, where separate encoders process inputs in isolation before concatenation. (2) Regarding the Physical Disconnect, researchers currently introduce auxiliary modules or rely on state forecasting to approximate dynamics. However, these approaches remain superficial: late fusion limits deep cross-modal reasoning, while dynamic prediction often mimics physics without understanding causality.

\noindent\textbf{Directions:} To bridge these gaps, the field must simultaneously advance toward \textit{Native Multimodal Architecture}. This means converting visual and physical data into tokens at the very beginning of training. By placing all modalities into the same language and shared space, the model does not need complex alignment steps. It can simply reason over all data types together, leading to a more natural and direct understanding of the physical world.
An important next step is to develop a hybrid \emph{Latent-Physics-Semantic World Model}.  Such a model would internally represent 3D geometry, physical dynamics, semantic attributes and affordances.  Given vision, optional depth/point-cloud/tactile input and a language instruction, the system encodes a unified world state, simulates candidate future states (e.g., object motion, contact, stability, affordance changes), and plans by reasoning jointly over semantics and physics.  This integration grounds high-level semantic intent in physics-aware simulation, helping to close the gap between semantic understanding, perception, and physical interaction.

\subsection{Instruction Following, Planning, and Robust Real-Time Execution}
\label{subsec:complex_task}
Fig.~\ref{fig:challenge_2} illustrates the four levels of this challenge, which are elaborated in detail below.

\input{imgs/challenge_2}
\subsubsection{Parsing Complex Instructions}
\label{sec:4.2.1}
Task instructions for VLA are often multimodal and underspecified, and failures in understanding propagate to perception, planning, and control.
We highlight two primary sources of difficulty:
(i) Open-ended, multimodal instruction forms. Instructions are no longer plain text, as they may mix language with images, scene cut-outs, internet photos, or hand-drawn sketches. 
(ii) Ambiguity and underspecification. Commands like ``help me'' or ``clean this up'' omit crucial task parameters (i.e., what, where, how, when).

\noindent\textbf{(1) Open-Ended Instruction.}
To handle open-ended, mixed-modality prompts, recent methods attempt to \emph{interleave images and text} into a single sequence, and use the same sequence modeling mechanism for understanding and control.
OE-VLA~\cite{oe_vla} adopts a shared visual encoder for all images and a text tokenizer for all text, converting them into token streams that are strictly interleaved to preserve the original instruction order.
Similarly, Interleave-VLA~\cite{interleave_vla} introduces special tags to its tokenizer, allowing image feature vectors to be seamlessly inserted within a text sequence. These approaches enable the policy to understand non-text instructions and improve direct cross-modal grounding without relying on standardized phrasing.

\noindent\textbf{(2) Ambiguous Instructions.}
Another line of work focuses on endowing the model with \emph{deeper reasoning and interactive clarification capabilities}. When facing ambiguous commands, ThinkAct~\cite{think_act} infers and verifies the intended target via scene parsing and feedback, while DeepThinkVLA~\cite{deepthink_vla} resolves ambiguity with causal chain-of-thought and aligns subgoals with correct execution through outcome-driven RL. 
When spatial information is underspecified, InSpire~\cite{inspire} explicitly prompts the policy to answer ``where is the target relative to the robot?" before acting, thereby auto-filling missing cues. Taking this a step further, AskToAct~\cite{asktoact} trains an ambiguity-recognition module on synthetically incomplete queries and uses large-scale clarification dialogues to teach the agent to proactively request missing details when an instruction is underspecified.

\subsubsection{Hierarchical Planning and Task Decomposition}
\label{sec:4.2.2}
While many VLA frameworks are optimized for short-horizon skills, executing long-horizon operations remains a largely unresolved challenge~\cite{long_vla}.
Agents must decompose high-level instructions into structured subgoals to act robustly.
Pure end-to-end models, which directly map inputs to low-level actions without explicit intermediate reasoning, often struggle with multi-step planning and compositional tasks~\cite{resilient_agile_mfg,thinkvla}.
To address this, hierarchical decomposition is a dominant paradigm~\cite{hamster, hirt, dual_process_vla, hume,gemini_robotics_15}.
Based on the type of intermediate representation they employ to bridge high-level intent and low-level control, current approaches can be broadly categorized into three families.

\noindent\textbf{(1) Language-Driven Planning.}
These methods adopt a \emph{modular hierarchical paradigm}, leveraging language to decompose tasks in semantic space. 
$\pi_{0.5}$~\cite{pi0_5} embeds hierarchical reasoning within a single inference chain: the model first proposes explicit language-level sub-tasks from vision and instructions, then conditions continuous control on these sub-tasks. 
OneTwoVLA~\cite{onetwo_vla} performs structured textual reasoning at key decision points, generating scene descriptions, high-level plans, and next-step instructions, to decompose tasks within the semantic space. Hi Robot~\cite{hirobo} employs a two-layer scheme where a VLM parses instructions into atomic sub-tasks, and a VLA controller handles low-level execution. Other methods use \emph{end-to-end hierarchical paradigm}, like
LoHoVLA~\cite{lohovla}, using a common VLM backbone to jointly produce language sub-steps and continuous actions, enabling long-horizon reasoning without a strict planner-executor split.

\noindent\textbf{(2) Planning via Multimodal Intermediates.}
These methods perform planning via multimodal intermediates, using non-linguistic representations like visual goals or affordances as the stepping stones for decomposition. 
On the \emph{vision-driven} side, CoT-VLA~\cite{cot_vla} employs pixel-level subgoal images as explicit intermediates~\cite{cot_vla}, while Embodied-SlotSSM~\cite{embodied_slotssm} employs slot-based~\cite{sssm} object-centric representations to create structured visual intermediates. HiP~\cite{hip} further extends this idea with a three-tier pipeline in which an LLM generates abstract subgoals, a video diffusion model produces physically feasible visual trajectories, and an inverse dynamics model converts these trajectories into actions.
On the \emph{affordance-driven} side,
RT-Affordance~\cite{rt_affordance} plans tasks by decomposing complex robotic manipulation into manageable affordance plans.
CoA-VLA~\cite{coa_vla} internalizes an affordance chain at each step as an implicit planning signal. 

\noindent\textbf{(3) Compositional Planning with Skill Libraries.}
These methods decompose long-horizon tasks into reusable atomic skills and compose them into higher-level behaviors for efficient and interpretable task execution.
For the \emph{explicit skill} usage, VLP~\cite{atomic_skill} builds a fine-grained library for data-efficient reuse of manipulation patterns.
Agentic Robot~\cite{agentic_robot} derives a short, semantically clear subgoal sequence from the library, decomposing a task into 2--5 verifiable atomic steps prior to execution.
RoboBrain~\cite{robobrain} also employs a hierarchical paradigm that expands human-understandable abstract instructions into executable atomic action sequences, achieving an intent–plan–action mapping through the joint learning of data and models. 
Other works explore the emergence of \emph{implicit skills}. For instance, DexVLA~\cite{dex_vla} learns to automatically annotate semantic sub-steps within long-horizon action sequences through temporal alignment.
AgiBot World~\cite{agitbot_colosseo} serves as a transition, using explicit skills during data collection but learning a policy that implicitly compresses high-dimensional control into semantic latent action tokens, enabling the emergence of composable behaviors.

\subsubsection{Error Detection and Autonomous Recovery}
\label{sec:4.2.3}
Long-horizon VLA deployments are inherently vulnerable to execution interruptions, perception drift, and actuation failures. Without timely, on-policy correction, small mistakes can compound into cascading failures that derail the entire task. To address this, research efforts have largely followed two main lines of inquiry:

\noindent\textbf{(1) Human-in–the-Loop Correction.}
These methods leverage a human user as an external source of intelligence to guide recovery. This can be \emph{reactive}, where the human provides corrective signals during execution. For instance, 
Yell At Your Robot~\cite{yell_at_your_robot} integrates real-time human language feedback as corrective signals for immediate behavioral adjustment, while
CLIP-RT~\cite{clip_rt} treats human language feedback as an ideal action template and embeds it into the decision process via similarity matching for efficient, retrain-free correction.
This approach can also be \emph{proactive}, where the agent solicits help when it detects ambiguity.
OneTwoVLA~\cite{onetwo_vla}, for example, incorporates active human clarification as a key component, proactively querying for user input to resolve uncertainty before acting.

\noindent\textbf{(2) Self-Correction.}
A more effective strategy is to enable the model to autonomously detect anomalous states and correct them. Specifically,
CorrectNav~\cite{correctnav} enables \emph{self-recovery without extra modules} by iteratively collecting the model’s own error trajectories, automatically identifying deviations, and generating corrective actions and visual data to continuously fine-tune the model.
Similarly, FPC-VLA~\cite{fpc_vla} uses a VLM to assess the semantic appropriateness of key actions and, when necessary, generates natural language feedback with corrective directions.
Agentic Robot~\cite{agentic_robot} focuses on the architectural level, which achieves autonomous correction via a \emph{standardized plan–act–verify closed loop}: a vision–language validator dynamically assesses subgoal completion and, upon failure, triggers predefined recovery strategies, effectively suppressing error accumulation.

\subsubsection{Real-Time Execution and Computing Efficiency}
\label{sec:4.2.4}
The powerful capabilities of VLA come at the cost of substantial computational overhead. Yet, physical-world interaction is highly sensitive to latency, especially in complex and long-horizon tasks. Bridging the compute-latency gap between model capability and real-time performance is thus critical to the practical deployment of VLA systems.
To address these issues, recent works focus on four directions:

\noindent\textbf{(1) Static Optimization of Architecture.}
A line of work focuses on static architectural optimization, which reduces inherent computational complexity by refining the model’s structure.
A common solution is \emph{compression} and \emph{quantization}.
BitVLA~\cite{bitvla} achieves ultra-low-precision efficiency via ternary 1-bit compression and distillation, while Evo-1~\cite{evo_1} offers a similar lightweight design with only 77M parameters.
SQAP-VLA~\cite{sqap_vla} introduces perceptual pruning strategies on the basis of quantization, and achieves a nearly two times inference speedup and half memory reduction.
Besides, some methods directly adopt \emph{lightweight backbones}, like NORA~\cite{nora} and TinyVLA~\cite{tinyvla}, while VLA-Adapter~\cite{vla_adapter} introduces lightweight adapters to graft knowledge from a large model onto a smaller policy network.
Other approaches fundamentally replace the computationally expensive Transformer attention mechanism with \emph{linear attention}.
SARA-RT~\cite{sara_rt} converts high-cost Transformer policies into linear-attention variants to cut inference delay. 
RoboMamba~\cite{robomamba} replaces the Transformer with the Mamba, attaining linear-time scaling and faster inference without explicit quantization or specialized accelerators.

\noindent\textbf{(2) Dynamic Optimization of Decoding Process and Inference Strategies.}
Beyond static architectural changes, this line focuses on runtime adaptivity, dynamically adjusting compute budgets during decoding and inference based on task complexity, thereby reducing latency and computation while maintaining accuracy. 
One strategy is to create \emph{dynamic inference paths}, which dynamically skip certain computation layers or terminate inference early at shallow depths, based on the complexity of the current input. For example,
MoLe-VLA~\cite{mole_vla} leverages layer skipping to reduce FLOPs, while CEED-VLA~\cite{ceed_vla} and DeeR-VLA~\cite{deer_vla} design early exit mechanisms. 
Another is to perform dynamic token processing through \emph{token pruning} or \emph{caching}.
VLA-Cache~\cite{vla_cache} designs adaptive caching strategies that treat static and dynamic tokens differently.
SpecPrune-VLA~\cite{specprune_vla} performs action-aware pruning conditioned on history and current observations.
CogVLA~\cite{cogvla} also reduces computation through instruction-driven visual token sparsification.
Furthermore, methods employ \emph{accelerated decoding} to overcome the sequential bottleneck of traditional approaches. For instance, Accelerating VLA~\cite{accelerating_vla} and OpenVLA-OFT~\cite{openvla_oft} generate an entire action chunk in a single forward pass through parallel decoding. Spec-VLA~\cite{spec_vla} adopts speculative decoding to emit candidate action tokens in a single forward pass with relaxed acceptance.

\noindent\textbf{(3) Optimization of Action Representation and Generation Paradigm.}
This type of method posits that the bottleneck in inference efficiency stems largely from how actions are represented and generated. By rethinking and optimizing action representations, efficiency can be fundamentally improved.
One strategy is \emph{efficient action tokenization}, which designs more compact and information-dense action tokens to reduce the number of prediction steps.
For example, FAST~\cite{fast} compresses action sequences to reduce training cost and wall time. XR-1~\cite{xr_1} leverages discrete visual–motor representations learned by VQ-VAE~\cite{vq_vae} to guide policy learning, while VQ-VLA~\cite{vq_vla} extends this idea by using a VQ-VAE tokenizer to compress long trajectories into a small set of discrete tokens.
Another strategy is \emph{asynchronous execution and inference}, where the system predicts the next action chunk while the current one is being executed, as seen in SmolVLA~\cite{smolvla} and Real-Time Action Chunking~\cite{rt_action_chunking}.
A third strategy focuses on \emph{accelerating diffusion policies} by reducing the number of required sampling iterations.
Time-Diffusion Policy~\cite{time_diffusion_policy} replaces the traditional time-varying denoising process with a fixed, direction-consistent unified velocity field.
Discrete Diffusion VLA~\cite{discrete_diff_vla} discretizes actions into tokens and employs masked diffusion with parallel prediction, alleviating the autoregressive decoding bottleneck.

\noindent\textbf{(4) Optimization of Training Paradigm and System.}
This kind of work emphasizes the design of the training process and the implementation of the system to further reduce the inference overhead and improve the execution efficiency.
A common principle among these approaches is to leverage additional knowledge or data during training so the model can \emph{take shortcuts at inference time}. For instance, ECoT-Lite~\cite{efficient_embodied_training} uses reasoning traces during training but completely bypasses explicit reasoning steps during inference. V-JEPA 2~\cite{v_jepa2} reduces planning overhead by predicting \emph{compressed semantic representations} instead of raw pixels. Meanwhile, Fast-in-Slow~\cite{fis} employs an elegant \emph{dual-system architecture within a single model}, enabling tight coordination between slow, deliberate reasoning and fast, reactive execution.
At the highest system level, some works elevate optimization to the \emph{operating system or distributed learning}. For example, AMS~\cite{os_action_primitives} introduces OS-level action context caching and replay mechanisms, and FedVLA~\cite{fed_vla} explores efficient distributed training of VLA models under a federated learning framework.

\subsubsection{Future Directions}
\noindent\textbf{Summary \& Trends}: 
To handle complex tasks, the community is currently divided into rigid hierarchical systems (using LLMs as high-level planners for code generation or sub-goal decomposition) for long-horizon reasoning, or massive end-to-end policies via instruction tuning for reactive skills. However, the former suffers from severe information loss between modules, while the latter lacks the reasoning capability for multi-stage correction, resulting in open-loop execution without introspection.

\noindent\textbf{Directions}: Future architectures must break this dichotomy by becoming \textit{Adaptive}. Just like a human, the model should decide how much to think based on the task. For simple tasks like grabbing a cup, it should act instantly. For complex tasks like assembling furniture, it should automatically activate deeper reasoning skills to plan steps. To do this, one direction is to use \textit{Unified Decision Tokens}. By treating seeing, thinking, and acting as a single stream of data, the model can naturally switch between fast action and deep thought without needing separate, rigid modules. This creates a true end-to-end unified mind that handles both simple reflexes and long-term planning.
Beyond just acting efficiently, robots need to change how they understand their own actions. Today's robots are passive, i.e., they just follow instructions without asking why. Future VLA models must evolve toward \textit{Self-Awareness}. The goal is an agent that not only knows what to do, but also understands why it is doing it. Models should shift from open-loop execution to closed-loop resilient autonomy, dynamically switching between replanning and reflex adjustment to autonomously recover from failures without intervention.

\subsection{From Generalization to Continuous Adaptation}
\label{sec:4.3}
\input{imgs/challenge_3}
Fig.~\ref{fig:challenge_3} illustrates the four levels of this challenge, which are elaborated in detail below.

\subsubsection{Open-World Generalization}
\label{sec:4.3.1}
Despite strong cross-modal understanding and manipulation in closed settings, large VLA models often generalize poorly when deployed in open, dynamic real-world environments. 
Conventional imitation learning relies on large human-annotated datasets and fails to cover the long tail of real scenes.
Therefore, achieving robust open-world generalization is a pivotal challenge.

\noindent\textbf{(1) Knowledge Transfer and Utilization.}
The most dominant approach posits that the key to generalization lies not in learning from scratch, but in effectively transferring vast prior knowledge from large-scale data sources. 
This is pursued in two main ways. \emph{Multi-task/multi-robot pretraining} involves training on massive robotic datasets to learn a general, hardware-agnostic prior over behaviors. For example, Octo~\cite{octo} pretrains a Transformer on about 800k robot trajectories to acquire general manipulation regularities and uses lightweight adapters for efficient fine-tuning, enabling rapid adaptation to new sensors and action spaces under limited data and compute. 
DexVLA~\cite{dex_vla} introduces billion-parameter diffusion action experts that pretrain across robot morphologies and adopts a three-stage curriculum to realize task-agnostic language–action mapping. 
RoboCat~\cite{robocat} pretrains on heterogeneous multi-robot data and continually improves on real trajectories for sustained task transfer. 
Dita~\cite{dita} leverages the large OXE dataset~\cite{oxe} and diffusion Transformers to learn cross-environment behaviors, adapting with as few as $10$ real demonstrations. 
EO-1~\cite{eo_1} further scales this paradigm by pretraining a shared backbone on the 1.5M-EO-Data dataset to achieve knowledge transfer and enhance open-world understanding.
The second method is \emph{internet/human video knowledge transfer}, which leverages data sources vastly larger than robotic datasets.
Following CLIP~\cite{clip}, R3M~\cite{r3m} extends this paradigm to robotics by pretraining visual encoders on massive collections of human first-person videos (e.g., Ego4D~\cite{ego4d}), thereby transferring general interaction knowledge into robotic policies.
In addition, the GR series (e.g., GR-1~\cite{gr_1}, GR-2~\cite{gr_2}) stands as a representative line of work in this direction that pre-train on massive human egocentric video datasets to transfer general physical and interaction knowledge into robotic policies.

\noindent\textbf{(2) Paradigm-Level Innovations.}
Beyond knowledge transfer from pretrained models or web-scale data, a growing body of work explores how models learn, not just what they learn, which is a key to achieving robust generalization.
For example, ICIL~\cite{icil_next_token} follows the \emph{in-context learning} paradigm that trains the model to infer tasks from a few demonstrations provided in the prompt at test time, enabling rapid, retrain-free adaptation.
Another direction focuses on \emph{emergent compositionality}, where methods like TRA~\cite{successor_features_tra} use a temporal contrastive loss to imbue the learned representation space with a compositional structure, allowing the model to automatically combine learned skills into new tasks. 
A more profound shift is toward \emph{conceptual generalization}, which moves beyond imitating actions to understanding semantic concepts. ObjectVLA~\cite{objectvla} jointly trains on robot trajectories and box-labeled VL corpora to achieve zero-shot manipulation of unseen objects, while LERF~\cite{lerf} fuses CLIP with 3D NeRFs for natural-language localization and grasping of novel objects.
Finally, to achieve robust deployment, new \emph{adaptation paradigms} emerge. Align-Then-Steer~\cite{align_then_steer} proposes a non-invasive adaptation method that steers a frozen VLA model's outputs using a lightweight, latent-space adapter. Robot Utility Models (RUM)~\cite{rum} pair large-scale home demonstrations with multimodal LLM reasoning for runtime verification and automatic retries, achieving zero-shot deployment in new environments.

\noindent\textbf{(3) Enhancing Data Diversity.}
Given the high cost of collecting real-world data, recent work expands the data distribution using generative models and semantic priors to build large-scale, more diverse training data at low robot cost. 
For \emph{data augmentation}, CACTI~\cite{cacti} scales multi-task imitation by using Stable Diffusion for zero-shot inpainting of expert images to increase layout and appearance diversity without additional robot rollouts. 
GenAug~\cite{genaug} employs text-to-image synthesis conditioned on a few demonstrations and prompts to produce visually diverse yet functionally consistent scenes, improving robustness to unseen environment shifts. For \emph{semantic augmentation}, ROSIE~\cite{sem_imagined_experience} distills knowledge from internet-scale VLMs into robot training, exposing policies to richer semantic combinations and task variants to strengthen open-set generalization.

\noindent\textbf{(4) Adaptive Architectural Design. }
Beyond the above approaches, the design of the model architecture itself profoundly influences its generalization capability. 
Specifically, \emph{hierarchical designs} enhance generalization by decomposing tasks into high-level planning and low-level execution. The high-level planner can leverage abstract knowledge learned from large-scale data, while the low-level executor focuses on acquiring reusable skills~\cite{lmm_3d_skill}. 
\emph{Multimodal fusion frameworks} that dynamically fuse multimodal sensor inputs can significantly enhance robustness in complex environments, like BAKU~\cite{baku}.
Meanwhile, \emph{generative diversity} methods like StructDiffusion~\cite{structdiffusion} use language-guided diffusion to generate multiple physically plausible action structures instead of a single deterministic plan, improving robustness to unseen object sizes and shapes.

\subsubsection{Continual Learning and Incremental Skill Acquisition}
\label{sec:4.3.2}
An embodied agent’s learning process should not end at deployment. It must continually acquire new skills throughout its lifetime to adapt to evolving environments and user needs.
However, recent studies reveal a critical issue: as new tasks are learned, the parameters supporting previously acquired skills are often overwritten, leading to sharp performance regressions and the erosion of multimodal reasoning capabilities inherited from backbones~\cite{speci,instruct_vla}.
To solve this, existing efforts broadly follow two routes.

\noindent\textbf{(1) Parameter Isolation and Expansion.}
These methods allocate dedicated parameter space for new skills or adopt modular designs that safeguard existing weights, thereby fundamentally preventing weight conflicts between old and new tasks and mitigating cross-task interference at its source. One prominent approach is \emph{Prompt-Based and Codebook-Based Learning}, which encodes skill knowledge into a set of discrete, composable prompts or codebook entries. When acquiring a new skill, the system simply adds a new prompt or codebook entry without modifying existing components~\cite{think_small_act_big,speci}. 
The other approach uses \emph{modular and expert-based architectures} to isolate knowledge. For example, InstructVLA~\cite{instruct_vla} adopts a two-stage training paradigm and a Mixture-of-Experts architecture to intelligently route between reasoning and action modules, avoiding direct modification of its backbone. Similarly, the scalable PerceiverIO proposed in iManip~\cite{iManip} falls into this category by adding new, skill-specific weights while freezing old ones.

\noindent\textbf{(2) Replay-based Knowledge Consolidation.}
Inspired by human review, these methods \emph{rehearse a subset of past examples} while learning new tasks to reinforce retained knowledge. Since storing and replaying all historical data is impractical, the core challenge lies in intelligently selecting the most informative samples for replay. ExpReS-VLA~\cite{expres_vla} addresses this by introducing compressed experience replay to mitigate catastrophic forgetting in robotic VLA systems, while iManip~\cite{iManip} proposes a temporal replay strategy that avoids random sampling and instead replays critical frames during skill execution.

\subsubsection{Sim-to-Real Gap in Deployment}
\label{sec:4.3.3}
The sim-to-real gap remains a core obstacle for deploying VLA policies, as discrepancies between simulated and real-world dynamics (e.g., friction, latency, actuation response) and perception (e.g., illumination, textures, sensor noise) severely degrade policy transfer despite the low-cost, large-scale data provided by simulators~\cite{slim}. To address this challenge, researchers have explored a variety of strategies:

\noindent\textbf{(1) Enhancing Simulation Fidelity and Robustness.}
The goal of this class of methods is to improve the direct transferability of policies, either by making the simulation environment more closely resemble the real world or by making the policy robust to the discrepancies between simulation and reality. 
A straightforward solution is to \emph{enhance the visual fidelity} of the simulator’s rendering. ManiSkill3~\cite{maniskill3} leverages GPU-parallel rendering, domain randomization, and background composition to narrow the appearance gap and enable zero-shot transfer. 
Another alternative to improving the simulation is to make the policy more robust by \emph{learning a stable intermediate representation}. SLIM~\cite{slim}, for instance, compresses high-dimensional RGB images into segmentation and depth maps, thereby filtering out task-irrelevant visual differences between sim and real.

\noindent\textbf{(2) Data-driven Simulators.}
Recognizing that classical physics engines cannot fully capture real-world complexity, a complementary line sidesteps explicit sim modeling by learning from or generating experiences using real-world data. 
One direction is \emph{generative augmentation} on real-world data, which attempts to expand a small set of real robot trajectories to enhance data diversity. For instance, GenAug~\cite{genaug} leverages web-scale image generative models to synthesize visually diverse but functionally consistent images from a few real robot demonstrations and semantic prompts, bypassing simulators entirely by exploiting the model’s prior over real-world visuals to generate highly realistic scenes. Another mainstream direction redefines physics-based simulation as \emph{data-driven prediction}: it trains a powerful world model to learn physical dynamics and causal relationships directly from massive amounts of real-world data, such as DreamGen~\cite{dreamgen}.
RynnVLA-001~\cite{rynnvla_001} further advances this direction through large-scale video generation pretraining combined with human-centric trajectory perception modeling, enabling implicit transfer of human manipulation skills to robotic control.

\subsubsection{Online Interaction and Reinforcement Learning}
\label{sec:4.3.4}
Imitation learning allows VLA models to quickly learn basic skills from offline data, but is limited by distributional shift and a performance ceiling capped by human demonstrators. Reinforcement Learning (RL) addresses these by enabling autonomous exploration, yet its application to large VLA models in high-dimensional continuous action spaces is hindered by low sample efficiency~\cite{improving_vla_online_rl,rldg,co_rft,refined_policy_distill} and the difficulty of designing effective rewards~\cite{vla_rl,grape,nora_1.5}.
To tackle these challenges, researchers integrate RL with VLA models’ strong priors, primarily through two directions:

\noindent\textbf{(1) Optimizing the Learning Process.}
Rather than letting RL explore from scratch, this approach injects or distills the rich knowledge and structural priors already learned by VLA models into the RL policy, addressing the slow and unstable nature of RL training. 
For \emph{knowledge transfer},
RLDG~\cite{rldg} first trains task-specialist RL policies, then distills their high-quality trajectories into a general VLA, improving precise control and generalization without fragile end-to-end RL fine-tuning. 
Refined Policy Distillation~\cite{refined_policy_distill} adds a simple MSE constraint so that VLA action distributions guide the RL agent, maintaining stability under sparse rewards and viewpoint changes. 
iRe-VLA~\cite{improving_vla_online_rl} alternates phases: it freezes the large backbone and trains a lightweight action head during RL for stability; it then unfreezes and fine-tunes with successful/expert trajectories under supervision to regain capacity. Beyond the above, some approaches \emph{optimize the internal structure} of RL algorithms. For example,
CO-RFT~\cite{co_rft} designs a chunked temporal-difference learning mechanism that feeds entire action sequences into the critic to predict multi-step returns, aligning with VLA’s chunked structure and significantly improving training stability and sample efficiency under limited data.

\noindent\textbf{(2) Automating Reward Generation.}
Instead of costly hand-crafted rewards or labor-intensive preference labels, recent work leverages VLM/LLM perception and reasoning to automatically derive dense, high-quality rewards directly from observations and goals.
One direction infers rewards through \emph{perceptual alignment} by measuring similarity between the current visual state and the goal description in a shared embedding space. VLM-RMs~\cite{vlm_zero_shot_reward} introduces this idea, and RoboCLIP~\cite{roboclip} extends it to video trajectories by computing video–language similarity for sparse rewards. Affordance-Guided RL~\cite{affordance_vl_prompt} converts VLM-predicted grasp points and target trajectories into continuous dense rewards that guide policy optimization.
A second direction uses VLMs as critics to \emph{rank trajectories or states} rather than relying on direct similarity scores. RL-VLM-F~\cite{rl_vlm_f} employs GPT-4V to compare observation pairs and infer preferences for training a reward function without human labels, while GRAPE~\cite{grape} decomposes tasks and generates stage-wise preferences for structured, multi-objective rewards.
A third direction leverages LLMs’ zero-shot \emph{code generation and high-level reasoning} to produce reward functions. Eureka~\cite{eureka} prompts an LLM with environment code and task specifications to generate executable rewards, VIP~\cite{vip} views reward learning as implicit value optimization from video, and VLA-RL~\cite{vla_rl} fine-tunes a VLM into a structured process–reward model that transforms sparse feedback into next-action-token supervision.

\subsubsection{Future Directions}

\noindent\textbf{Summary \& Trends}: 
To achieve generalization, the dominant approach currently hinges on Scaling Laws, i.e., aggregating massive, heterogeneous datasets to train large-scale transformers via passive imitation learning. While this has significantly improved task-level success rates on seen distributions, models remain hardware-dependent and temporally static. They are frozen after training, lacking the agency to actively explore or adapt to novel robot morphologies without extensive fine-tuning.

\noindent\textbf{Directions}: To realize ``GPT moment" in embodied intelligence, the paradigm must shift from training fragmented, robot-specific policies toward developing \emph{Morphology-Agnostic Representations}. By logically disentangling high-level semantic planning from low-level proprioceptive control, a unified brain can transfer manipulation skills across vastly different embodiments—from quadrupeds to humanoids—via lightweight, modular adapters. This would enable true \emph{Zero-Shot Cross-Embodiment Transfer}, where a new robot is treated simply as a new peripheral for a universal policy.
Furthermore, this generalization must be sustained through time via \emph{Autonomous Open-Ended Evolution}. We envision a shift from static training sets to a self-reinforcing data engine, where agents exhibit intrinsic motivation to act as curious explorers. By combining self-supervised exploration with online reinforcement learning, future VLAs will transition from passive imitators to active learners, identifying their own knowledge gaps and generating high-quality training data in the wild. This creates a virtuous closed loop of ``\emph{Deployment} $\rightarrow$ \emph{Discovery} $\rightarrow$ \emph{Evolution}," allowing the system to continuously refine its world model and expand its capabilities without human.

\subsection{Safety, Interpretability and Reliable Interaction}
\label{sec:4.4}
Fig.~\ref{fig:challenge_4} illustrates the two levels of this challenge, which are elaborated in detail below.
\input{imgs/challenge_4}

\subsubsection{Reliability and Safety Assurance}
\label{sec:4.4.1}
VLA models, particularly end-to-end deep learning systems, usually lack transparency in their decision-making and exhibit unpredictable behavior. When deployed in unstructured, human-shared physical environments, they may execute hazardous actions due to perception errors, generalization failures, or misinterpretation of instructions, potentially endangering humans, the environment, or themselves. Consequently, establishing a reliable and verifiable safety assurance mechanism is a critical prerequisite for the real-world deployment of VLA systems. To address this challenge, two directions are explored:
 
\noindent\textbf{(1) Constraint-Based Safety Paradigms.}
This paradigm injects explicit rule systems, inside or outside the model, to hard-bound the action space and avoid unsafe behaviors. 
Specifically, applying \emph{rule-based explicit constraints} is the most straightforward approach. AutoRT~\cite{autort} introduces a robot constitution via structured prompting to encode multi-level constraints for behavior bounding in the wild. 
Alternatively, some works directly \emph{internalize safety constraints} as an integral part of the model’s learning process. SafeVLA~\cite{safevla} explicitly models physically hazardous behaviors as a cost function within a constrained Markov decision process, where the training objective is to maximize task reward while ensuring the cumulative cost remains below a predefined safety threshold.

\noindent\textbf{(2) Learning-based Alignment Paradigms.}
\label{sec:4.4.2}
Since scenarios in the real world are highly complex and cannot be fully covered by a finite set of handcrafted rules, some methods aim to \emph{internalize a human-aligned safety intuition and judgment}, enabling models to proactively detect and avoid risks.
For example, Gemini Robotics~\cite{gemini_robotics} applies Constitutional AI post-training on safety data, ensuring that policies follow human-centric principles and thereby internalize safety intuition. 
Beyond passively adhering to predefined rules, the model must \emph{actively assess the uncertainty and potential risks} of the current situation and adapt its behavior accordingly. GPI~\cite{resilient_agile_mfg} integrates confidence estimation, probabilistic action generation, and language-guided backtracking to pause, seek help, or replan under uncertainty.
Furthermore, RationalVLA~\cite{rationalvla} introduces a learnable refusal token to reject unsafe/invalid commands, adding a rational safety layer between high-level semantics and low-level control.

\subsubsection{Interpretability and Trustworthy Interaction}
\label{sec:4.4.2}
Most VLA models follow the end-to-end deep learning paradigm, which is inherently a black box and offers little mechanistic insight~\cite{mechanistic}.
When a robot acts, its inability to explain its rationale to the user impedes debugging, erodes trust, and hinders efficient human-robot collaboration. Thus, a core challenge for VLA systems is to make decision logic more transparent and behavior more predictable. Research efforts are therefore shifting toward two aspects:

\noindent\textbf{(1) Enhancing Process Interpretability.}
The aim is to expose the model’s abstract neural states as explicit, human-understandable intermediates at each step of the think–decide–act chain.
\emph{Chain-of-thought reasoning} is a well-known approach to enhancing interpretability. The intermediate reasoning process can be expressed either in linguistic form or in visual form. For the former, 
Diffusion-VLA~\cite{diffusion_vla} conditions a diffusion policy on natural-language reasoning, exposing step-wise intent. ECoT~\cite{embodied_cot} outputs editable step-by-step rationales that users can correct via language. For the latter, CoT-VLA~\cite{cot_vla} adds visual subgoal images to render intermediate plans observable.
Moreover, in hierarchical architectures, the \emph{intermediate instructions generated by the high-level planner} inherently serve as a natural source of interpretability. 
For example, RT-H~\cite{rt_h} separates language–action generation from execution, enabling self-explanation and language-level intervention.
HiRobot~\cite{hirobo} outputs readable low-level commands from a high-level planner, making task decomposition transparent. GraSP-VLA~\cite{grasp_vla} explicitly converts visual inputs into symbolic states and performs planning in this symbolic space, making its intermediate process inherently interpretable.
Besides, recent efforts aim to \emph{decode the internal, hidden symbolic states} from trained, black-box VLA models. A representative work is DIARC-OpenVLA~\cite{probe_symbolic_vla}, which trains linear probes on hidden layers to explicitly map neural activations to symbolic states, providing a monitorable layer of decision transparency without altering the original model.

\noindent\textbf{(2) Behavioral Predictability.}
Beyond explaining why a decision is made, it is equally important to design robot behaviors that are \emph{inherently intuitive and aligned with human expectations}, thereby fostering trust directly through interaction. CrayonRobo~\cite{crayonrobo} externalizes the model’s internal decision logic using structured, semantically explicit visual prompts, creating a shared, interpretable language that lets humans intuitively understand and even design the prompts for deeper collaboration. 
Another critical aspect is \emph{predictable responses to dynamic instructions}.
SwitchVLA~\cite{switchvla} introduces structured task switching: upon mid-execution instruction changes, the agent rolls back conflicting actions before smoothly transitioning to the new goal, yielding natural, predictable behavior in open-ended interaction.

\subsubsection{Future Directions}
\noindent\textbf{Summary \& Trends}: 
Currently, safety is predominantly handled by extrinsic guardrails (e.g., rule-based shields or constitution-based filtering like AutoRT) or post-hoc rationalization (prompting VLMs to caption their actions). While providing a layer of protection, these reactive measures are separated from the policy's core decision process, often failing to prevent intrinsic model hallucinations or confident but wrong actions in real-time.

\noindent\textbf{Direction}: To build truly trustworthy embodied agents, the field must evolve beyond imposing static safety rules toward cultivating \emph{Intrinsic Uncertainty Awareness}. In unstructured open worlds, absolute safety cannot be guaranteed by pre-defined constraints alone. Instead, future VLA models require a System 2 reflective layer that actively estimates epistemic uncertainty, endowing the agent with a sense of doubt. This enables a paradigm shift from reactive emergency stops to \emph{Proactive Risk Aversion}, where the agent autonomously pauses to solicit human clarification or replans when it detects ambiguity or potential hazard.
Furthermore, trust relies on establishing \emph{Shared Mental Model} through intervention-ready transparency. Interpretability should not be merely a post-hoc debugging tool, but an integral part of the execution loop. We envision agents that visualize their thought process, such as future trajectories, attention heatmaps, or subgoal decompositions, before physical action is taken. Crucially, this transparency must be actionable: it should empower users to not only anticipate robot behavior but also intervene effectively. By allowing humans to correct the robot’s reasoning chain via natural language or gestures, we can close the loop of \emph{Interactive Safety}, ensuring that VLA systems are not just compliant, but genuinely aligned with human intent.

\subsection{Data Construction and Benchmarking Standards}
\label{sec:4.5}
Fig.~\ref{fig:challenge_5} illustrates the two levels of this challenge, which are elaborated in detail below.

\input{imgs/challenge_5}
\subsubsection{Multi-Source Heterogeneous Data}
\label{sec:4.5.1}
The capabilities and generalization of VLA models are fundamentally constrained by the scale, diversity, and quality of their training data. However, acquiring and unifying high-quality, large-scale, and diverse data presents a formidable challenge, primarily due to the inherent heterogeneity of data sources (e.g., sim vs. real, different robot embodiments) and their respective control interfaces. To address this, the research community systematically initiates explorations across three interconnected levels:

\noindent\textbf{(1) Representation-Level Unification and Alignment.}  
The core idea here is to model heterogeneous data within a shared, semantically consistent latent space, thereby eliminating heterogeneity at the cognitive level rather than directly handling raw discrepancies. This is achieved through two complementary strategies.
The first aligns behaviors in a latent action space by \emph{learning a unified discrete representation} that maps continuous, high-dimensional motions from different robots or human videos into semantically consistent action tokens. This filters out low-level control differences and aligns behaviors at a higher semantic level. LAPA~\cite{latent_action_pretrain}, Moto~\cite{moto}, and UniVLA~\cite{univla} learn such task-centric latent action representations through unsupervised or self-supervised video learning.
A more holistic strategy constructs a \emph{shared semantic space across all modalities and embodiments}, extending beyond action correspondence to unify perception, reasoning, and control. 
RDT-1B~\cite{rdt1b} and AgiBot World~\cite{agitbot_colosseo} map diverse robot actions into unified physical or latent vectors, while Scaling Cross-Embodied Learning~\cite{scaling_cross_embodied} tokenizes heterogeneous visual and proprioceptive inputs for a shared Transformer to handle multiple morphologies. At the multimodal level, methods such as RT-1~\cite{rt_1}, GR-2~\cite{gr_2}, ViSA-Flow~\cite{visaflow}, and Humanoid-VLA~\cite{humanoid_vla} achieve consistent VLA grounding through unified tokenization, semantic alignment, or self-supervised learning. Human-to-robot transfer approaches, including EgoVLA~\cite{ego_vla} and DexWild~\cite{dexwild}, further align human and robot motion using MANO hand models and inverse kinematics, enabling cross-domain embodied transfer.

\noindent\textbf{(2) Data-Level Augmentation and Optimization.} 
Rather than altering the model’s latent space, this line of work directly operates on raw data. 
The first strategy, \emph{generative data augmentation}, creates expanded data distributions using large pretrained generative models. This substantially increases visual diversity at low cost and improves robustness to appearance variations in heterogeneous real-world data. CACTI~\cite{cacti} and GenAug~\cite{genaug} augment robot data via inpainting or restyling, ROSIE~\cite{sem_imagined_experience} enriches data at the semantic level using VLM priors, and Models with Data Generation via Residual RL~\cite{modeling_data_with_rl} generate additional samples through RL to further strengthen downstream VLA performance.
The second strategy, \emph{automated data mixture optimization}, focuses on making better use of existing heterogeneous datasets by treating data fusion as an optimization problem.
Re-Mix~\cite{remix} adjusts sampling weights of heterogeneous data subsets based on performance feedback, enabling the model to focus on informative samples and achieve efficient cross-domain fusion.

\noindent\textbf{(3) Standardization and Benchmark Construction.}
This line of work reduces the heterogeneity of the data at the source by establishing standardized data collection protocols, synchronization mechanisms, and unified benchmarks.
A major focus is \emph{unified acquisition and synchronization} within individual datasets to ensure high quality and internal consistency. 
RH20T~\cite{rh20t} enforces strict temporal alignment across multimodal sensors, and BridgeData V2~\cite{bridgedata_v2} organizes diverse data types into a standardized format. In simulation, RoboCasa~\cite{robocasa} and CoVLA~\cite{covla} provide large-scale, high-fidelity environments that act as standardized digital laboratories.
Another effort involves \emph{collecting and aligning human-centric and multi-view data}, which is necessary for robots operating in human environments. Representative examples include Ego4D~\cite{ego4d} and EPIC-KITCHENS~\cite{epic_kitchens}, with Ego-Exo4D~\cite{egoexo4d} further integrating first- and third-person viewpoints to support learning skilled activities from multiple perspectives.
The broader ambition is \emph{cross-domain standardization}, where heterogeneous datasets are aligned at scale to form unified fusion benchmarks. Open X-Embodiment (OXE)~\cite{oxe} marks a major milestone by aggregating dozens of datasets into a single benchmark for cross-embodiment generalization. RoboMM~\cite{robomm} advances it through a three-level semantic alignment framework that enables joint training across multiple datasets.

% 5.2
\subsubsection{Evaluation Benchmarks}
\label{sec:4.5.2}
Standardized benchmarks play a pivotal role in embodied intelligence by establishing common evaluation protocols that enable fair comparison and reproducible research. However, as VLA models advance rapidly, the yardsticks used to measure them struggle to keep pace, revealing several critical limitations~\cite{experiences}. First, a lack of unified standards in metrics and experimental setups makes fair comparison difficult. Second, many existing benchmarks are limited to simple, short-horizon tasks, failing to test advanced cognitive reasoning. Third, they often lack a systematic way to probe frontier generalization capabilities. To address these gaps, the community actively develops a new generation of benchmarks and evaluations.

A primary direction of this effort is the pursuit of \emph{comprehensiveness and standardization}. The work on Benchmarking VLAs~\cite{benchmarking_vla} provides a blueprint by emphasizing unified I/O, metrics, and multi-robot coverage, shifting the focus from tasks to metrics. EUQ~\cite{vla_uncertainty_quality} introduces a human-assessed, multi-dimensional scoring system to capture process quality beyond binary success. At the infrastructure level, simulation platforms like ManiSkill3~\cite{maniskill3} and robosuite~\cite{robosuite} contribute standardized APIs and task suites, providing a reproducible foundation for fair and scalable evaluation.
A second major direction focuses on \emph{expanding the breadth and depth of tasks} to assess more complex capabilities. CALVIN~\cite{calvin} is designed to require the execution of long-horizon sequences of language-guided operations. LIBERO~\cite{libero} is introduced as the first benchmark specifically for lifelong learning in robotics, featuring standardized metrics for knowledge transfer and forgetting. Furthermore, Ego-Exo4D~\cite{egoexo4d} pioneers the synchronization of first- and third-person recordings for multi-perspective skill analysis.
Finally, a third direction aims to design more challenging tests that focus on frontier \emph{generalization and reasoning capabilities}. The From Intention to Execution~\cite{intention_to_execution} suite is introduced to probe the intention-execution gap and systematically covers challenges in object diversity, linguistic complexity, and visual-language reasoning. To specifically assess the abilities of instruction-tuned models, InstructVLA~\cite{instruct_vla} releases the SimplerEnv-Instruct benchmark, a comprehensive suite of 80 zero-shot tasks featuring multilingual expressions, novel objects, and implicit intentions to evaluate contextual reasoning and generalization.

\subsubsection{Future Directions}
\noindent\textbf{Summary \& Trends}: Driven by the pursuit of scaling laws, the field is currently preoccupied with aggregating massive, heterogeneous real-world datasets to fuel models. On the evaluation front, the standard remains simplistic, relying heavily on binary success rates in controlled settings. However, real-world collection is inherently unscalable and noisy, and binary metrics fail to capture the nuances of robustness, often masking critical failure modes.

\noindent\textbf{Directions}: To scale embodied intelligence, the field must transition towards a \emph{Simulation-First, Failure-Centric Paradigm}. Relying solely on real-world data is unscalable; instead, we envision \emph{Simulated Universes} acting as infinite data factories that generate diverse, labeled trajectories with perfect ground truth. The core challenge will be bridging the Sim-to-Real gap for perception and physics, allowing real-world data to serve efficiently as a high-quality alignment set to calibrate the simulator's physics and rendering fidelity, rather than being the primary training source.
Equally important is a shift in how we treat errors. Current pipelines often discard failed trajectories, wasting critical information. Future systems must \emph{Turn Failure into Signal}, treating mistakes as gold mines for negative mining and contrastive learning. By explicitly training on what not to do and diagnosing why failures occur, agents can learn not only to avoid risks but also to autonomously recover from inevitable execution errors. 
Finally, evaluation must evolve from simple binary success rates to \emph{Comprehensive Diagnostic Stress Testing}. Benchmarks should utilize high-fidelity simulation proxies to assess holistic capabilities—quantifying not only task completion but also safety margins, efficiency, and resilience to perturbations—thereby prioritizing robust adaptability over rote execution of memorized trajectories.

%% file: imgs/challenges_overall.tex
\begin{figure}[t]
  \centering
  \includegraphics[width=0.5\textwidth]{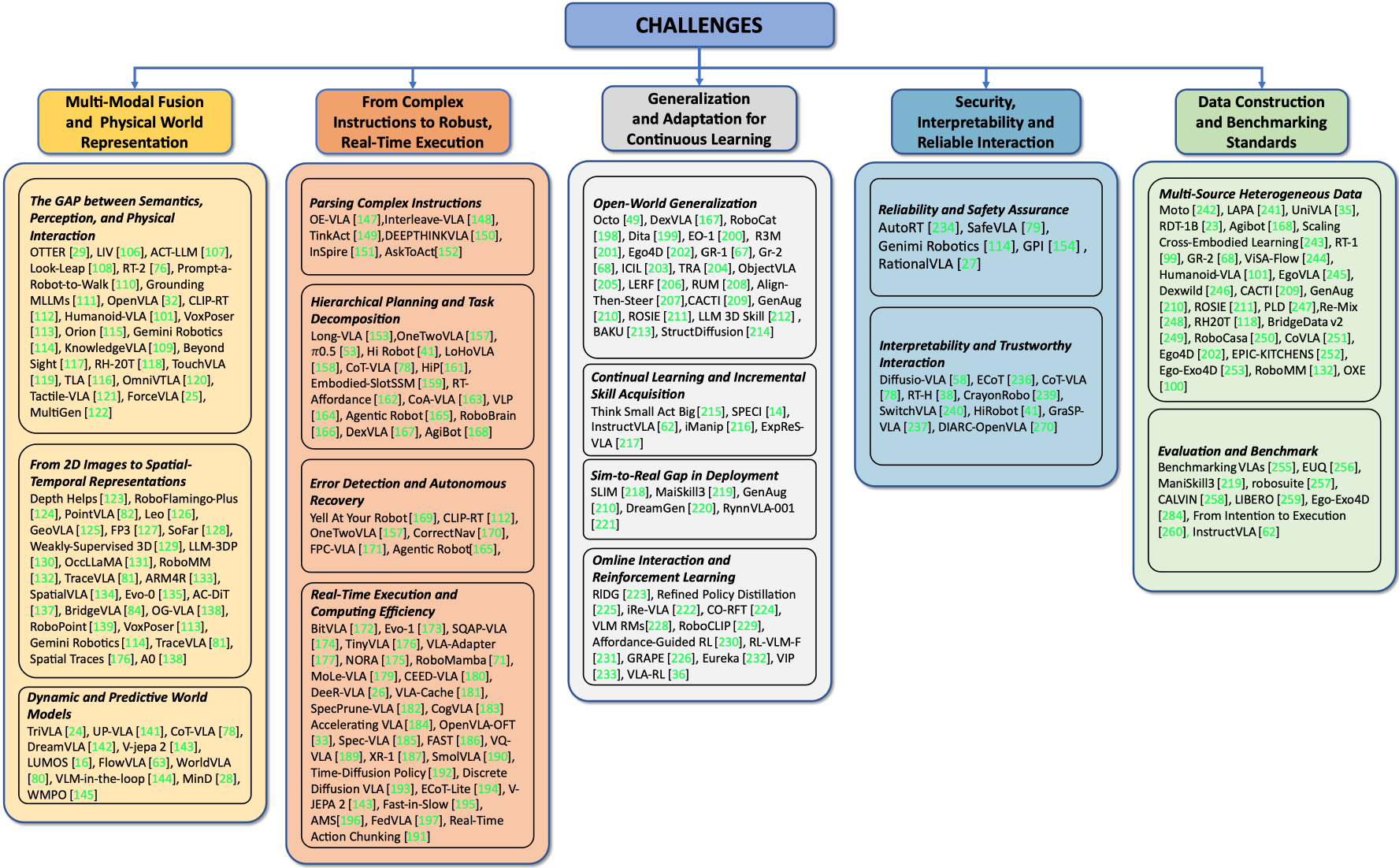}
  \caption{\textbf{Taxonomy of VLA challenges}, encompassing 5 primary challenges and 15 sub-challenges, with representative works listed. Please zoom in for more details.}
  \label{fig:Challenges_overview} 
\end{figure}

%% file: imgs/challenge_1.tex
\begin{figure}[!t]
  \centering
  \includegraphics[width=0.50\textwidth]{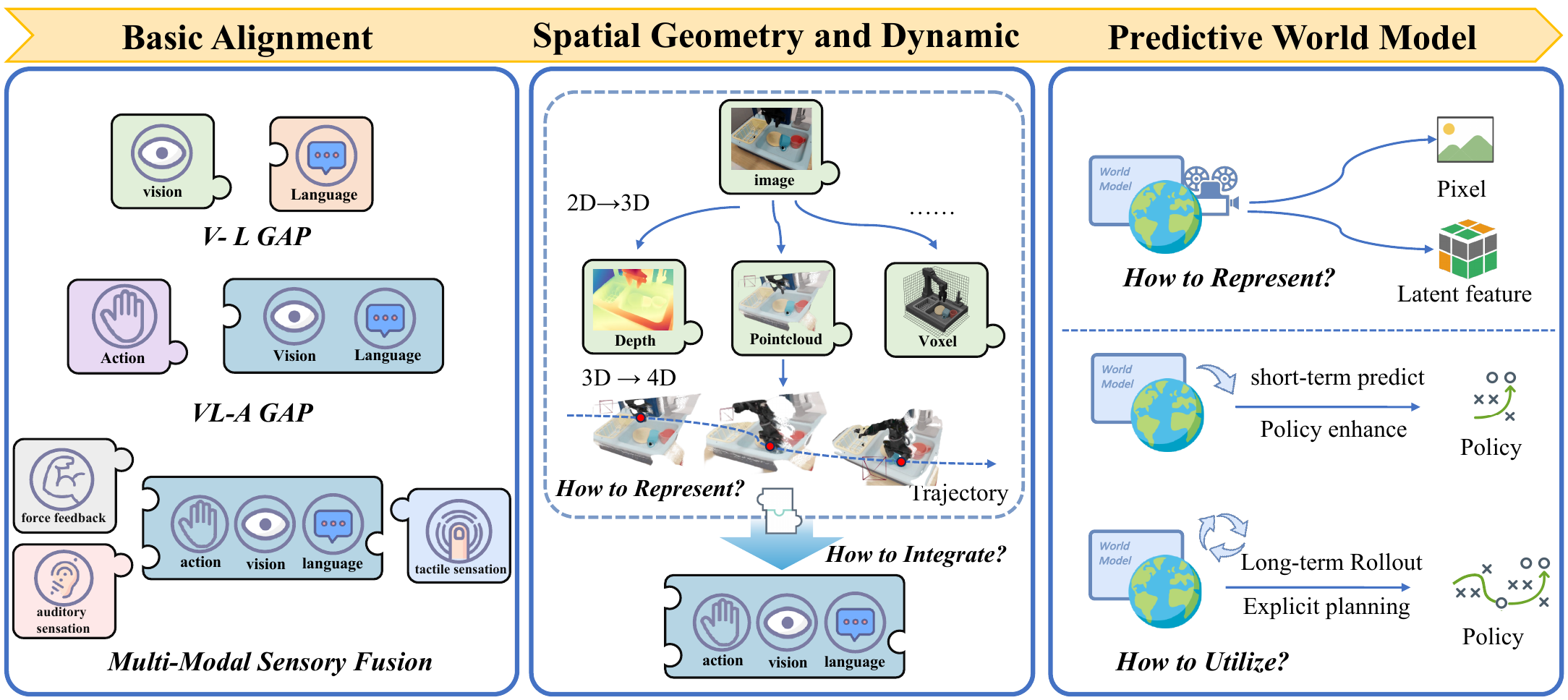}
  \caption{\textbf{The challenge of Multi-Modal Alignment and Physical World Modeling}. First, Section~\ref{sec:4.1.1} addresses the fundamental disalignment at the interface of information. Building upon this, Section~\ref{sec:4.1.2} focuses on the construction of the world's geometric and dynamic structure. Section~\ref{sec:4.1.3} represents the highest level of understanding, as embodied in dynamic predictive capabilities.
  }
  \label{fig:challenge_1}
\end{figure}

%% file: imgs/challenge_2.tex
\begin{figure*}[!t]
  \centering
  \includegraphics[width=0.85\textwidth]{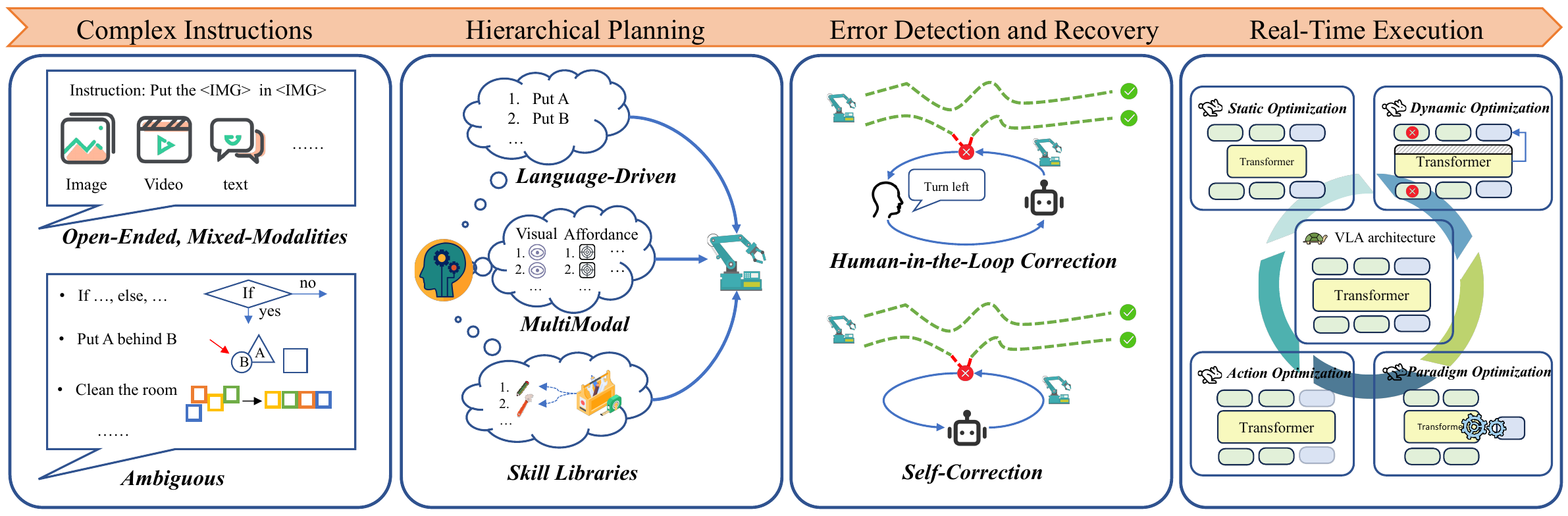}
  \caption{\textbf{The Challenge of Instruction Following, Planning, and Robust Real-Time Execution}. The flow begins with Section~\ref{sec:4.2.1}, where the model understands what a human wants, even if the instructions are unclear or mixed with images. This understanding then moves to Section~\ref{sec:4.2.2}, where big goals are broken down into smaller, workable steps or plans. For the third step, Section~\ref{sec:4.2.3}, the robot carries out its plan, watches for problems, and fixes them if something goes wrong. Lastly, Section~\ref{sec:4.2.4} is a rule for the whole process, requiring every step to happen quickly. 
  }
  \label{fig:challenge_2}
\end{figure*}

%% file: imgs/challenge_3.tex
\begin{figure*}[htbp]
    \centering
    \includegraphics[width=0.85\textwidth]{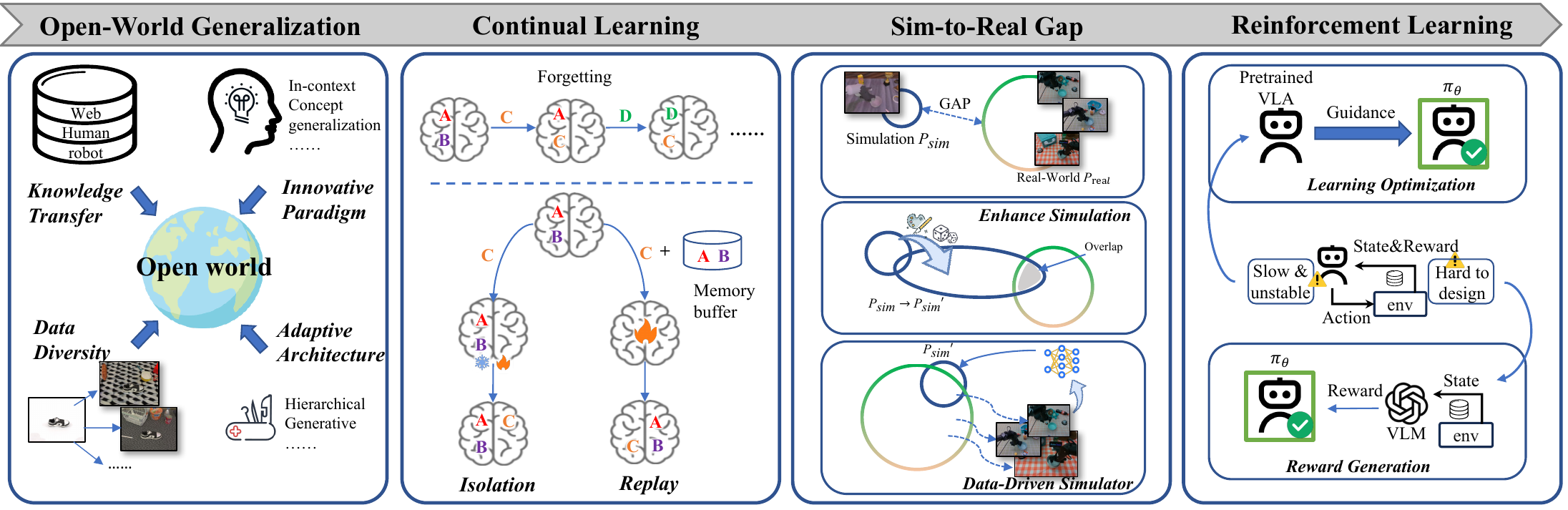}
    \caption{\textbf{The Challenge of From Generalization to Continuous Adaptation}. This diagram illustrates how VLA models operate continuously in dynamic, open-world environments, highlighting four key enabling strategies. Section~\ref{sec:4.3.1} represents the initial ability to perform well in settings not seen during training. Building on this, Section~\ref{sec:4.3.2} focuses on how agents can continuously acquire new skills throughout their operational lifetime without forgetting old ones. Section~\ref{sec:4.3.3} addresses the critical challenge of transferring learned policies from virtual environments to the physical world. Finally, Section~\ref{sec:4.3.4} highlights how agents refine their behaviors and learn from real-time experience. 
    }
    \label{fig:challenge_3}
\end{figure*}

%% file: imgs/challenge_4.tex
\begin{figure}[htbp]
    \centering
    \includegraphics[width=\columnwidth]{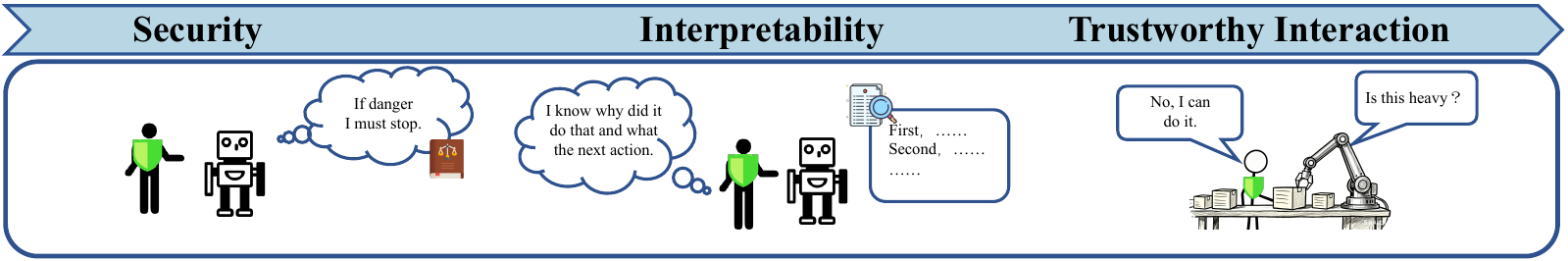}
    \caption{\textbf{The challenge of Safety, Interpretability and Reliable Interaction}. This diagram shows how VLA systems build human trust, broken into three key layers. Section~\ref{sec:4.4.1} is about making sure the robot is physically safe and works reliably. Moving up, Section~\ref{sec:4.4.2} focuses on helping humans understand why the robot makes certain decisions, making the robot's actions easy to understand and predict, leading to smooth collaboration. 
    }
    \label{fig:challenge_4}
\end{figure}

%% file: imgs/challenge_5.tex
\begin{figure}[htbp]
    \centering
    \includegraphics[width=\columnwidth]{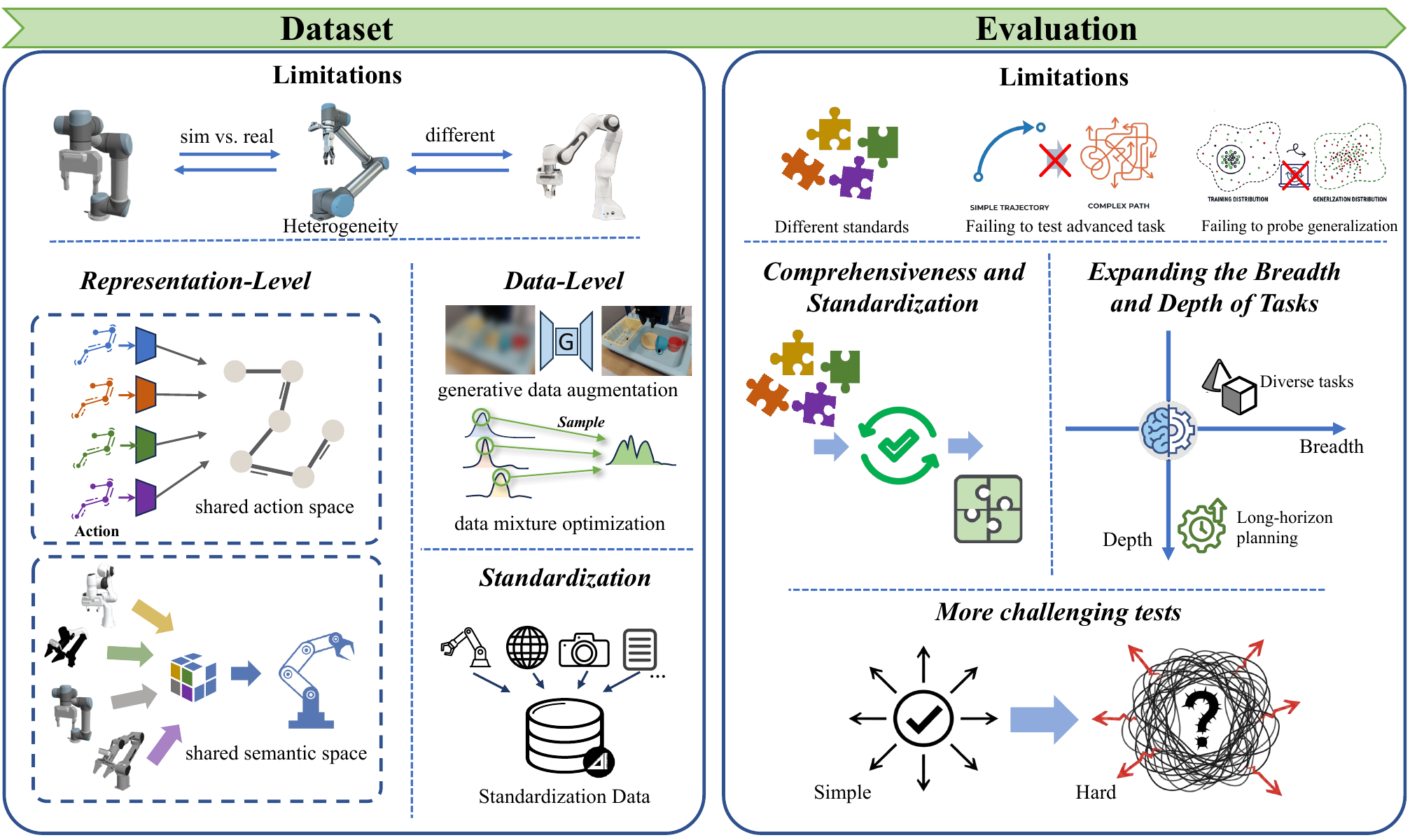}
    \caption{\textbf{The challenge of Data Construction and Benchmarking Standards}. Section~\ref{sec:4.5.1} addresses the critical bottleneck of acquiring and unifying diverse training resources to construct large-scale datasets. Section~\ref{sec:4.5.2} focuses on the standardization and increasing complexity of assessment protocols. 
    }
    \label{fig:challenge_5}
\end{figure}

%% file: sections/7_conclusion.tex
\section{conclusion}
This survey presents a comprehensive anatomy of Vision-Language-Action (VLA) models, structured to guide readers from basic modules and historical milestones to the core challenges at the research frontier. We provide a detailed analysis of the five key problem areas: representation, execution, generalization, safety, and dataset evaluation, reviewing current solutions, and highlighting future opportunities for each. We hope that this work serves as a foundational roadmap, helping both newcomers and experienced researchers navigate and advance the rapidly evolving field of embodied intelligence.

%% file: sections/X_supp.tex
\clearpage
\setcounter{page}{1}
\setcounter{figure}{0}
\setcounter{table}{0}
\setcounter{section}{0}

\renewcommand{\thefigure}{S\arabic{figure}}
\renewcommand{\thetable}{S\arabic{table}}
\renewcommand{\thesection}{\Alph{section}} 
\renewcommand{\thesubsection}{\Alph{section}\arabic{subsection}}

\section{Appendix}

\subsection{Applications}
\label{sec:5}
The true measure of Vision-Language-Action (VLA) models lies in their ability to solve real-world problems. By integrating perception, reasoning, and control, these models are uniquely equipped to translate abstract human intent into grounded, executable actions, bridging the long-standing gap between high-level cognition and low-level robotics. 
Leveraging large-scale pretraining, VLA-driven robots demonstrate unprecedented capabilities in generalization and adaptation, surpassing conventional modular pipelines in both autonomy and efficiency. This section surveys their transformative impact across two primary domains: household robotics and industrial automation.

\subsubsection{Embodied Manipulation and Household Robotics}
\label{sec:5.1}
The unstructured, dynamic, and human-centric nature of household environments makes them a major proving ground for VLA models. Unlike structured factory settings, homes require robots to understand natural language, handle a vast diversity of unseen objects, and perform complex, long-horizon tasks.

VLA models excel in this domain precisely because their core architecture is well-suited to these challenges. Their ability to leverage internet-scale knowledge allows them to recognize and interact with a near-infinite variety of household items without task-specific training (e.g., SayCan~\cite{saycan}, RT-1~\cite{rt_1}, RT-2~\cite{rt_2}). The evolution of benchmarks from ALFRED~\cite{alfred} to real-world validations like ChatVLA~\cite{chatvla} confirms their robustness in sequential reasoning. Furthermore, the hierarchical reasoning inherent in many VLA systems (e.g., Helix~\cite{helix}) enables the decomposition of vague commands like ``clean the kitchen" into concrete, executable subtasks.

Looking ahead, the next frontier for household VLA systems lies in achieving true personalization and collaborative intelligence. Future models must move beyond simply executing one-off commands and learn to understand a user's long-term preferences, habits, and implicit intentions. This requires a deeper integration of interactive learning, where robots can learn from real-time verbal feedback, ask clarifying questions when faced with ambiguity, and even proactively suggest actions based on learned routines. 

Furthermore, to become truly ubiquitous, these systems must operate on low-power, on-device hardware, necessitating breakthroughs in model efficiency and compression. The ultimate goal is to transform household robots from simple instruction-followers into proactive, adaptive, and truly personalized domestic assistants.

\subsubsection{Industrial and Field Robotics}
\label{sec:5.2}
Following their success in household scenarios, VLA models now extend to industrial domains, where they promise to bring unprecedented flexibility to manufacturing, logistics, and field operations. Industrial environments, however, impose far stricter demands on precision, reliability, and safety. The evolution of VLA models for industrial use is therefore characterized by a clear focus on enhancing their physical intelligence and robustness.

This evolution is proceeding along three major directions: (1) Incorporating physical perception via tactile and force sensors (e.g., Tactile-VLA~\cite{tactile_vla}, VTLA~\cite{vtla}); (2) Developing industrial-grade Reasoning for complex processes (e.g., ForceVLA~\cite{forcevla}, CogACT~\cite{cogact}
); (3) Ensuring Safety and Reliability through mechanisms like safe reinforcement learning (e.g., SafeVLA~\cite{safevla}).

The future of VLA in industry hinges on bridging the gap between flexible intelligence and the rigorous demands of production environments. The next wave of innovation will likely focus on certification-ready safety and formal verification, moving beyond empirical safety to provide provable guarantees on robot behavior. Another critical direction is zero-shot adaptation to new tasks and parts in highly customized manufacturing, where reprogramming a robot for every new product is economically infeasible. This requires VLA models to learn from CAD files, technical manuals, and video demonstrations of human workers. Finally, the integration of VLA into multi-agent systems enables fleets of robots to collaboratively perform complex assembly or logistics tasks, coordinated by a central language-based understanding of the overall production goal.

\subsection{Basic Modules}
\subsubsection{Training Strategy}
\label{sec:2.5}
Current VLA training follows three largely complementary routes that are often combined in practice:

\noindent\textbf{(1) Behavioral Cloning (BC).}
In current VLA research, BC is the dominant paradigm: it formulates control as supervised imitation, learning a mapping from multimodal observations (i.e., vision, language, proprioception) to expert actions by minimizing the prediction demonstration discrepancy.
In practice, BC underpins a broad spectrum of VLA systems across architectures, from diffusion-based controllers to multimodal Transformer generalists (e.g., Diffusion Policy~\cite{diffusion_policy}, TriVLA~\cite{trivla}, VIMA~\cite{vima}, Octo~\cite{octo}, RDT-1B~\cite{rdt1b}, RT-H~\cite{rt_h}, Hi Robot~\cite{hirobo}, GR-2~\cite{gr_2}, 3D-VLA~\cite{3d_vla}, RoboMM~\cite{robomm}). 
Beyond flat policies, BC is also employed as a pre- or post-training stage in hierarchical pipelines, e.g., adapting continuous control in $\pi_{0.5}$~\cite{pi0_5} and driving the fast S-Sys1 executor in Dual-Process VLA~\cite{dual_process_vla}.

\noindent\textbf{(2) Predictive Modeling.}
Instead of imitating actions, predictive modeling learns to anticipate the world in future observations or latent dynamics, providing powerful self-supervised signals that internalize physics and causality. 
World models exemplify this idea: WorldVLA~\cite{worldvla}, LUMOS~\cite{lumos}, and UVA~\cite{uvam} train with predictive and self-supervised objectives, enabling effective learning from unstructured data and strong performance on complex, long-horizon robotic tasks. 
Additionally, other approaches learn discrete latent actions from unlabeled video as in self-supervised prediction like world modeling (e.g., LAPA~\cite{lapa}).

\noindent\textbf{(3) Reinforcement Learning (RL).}
RL moves beyond demonstrations to optimize policies through interaction and reward feedback, and in VLA it is often built upon BC-pretrained backbones to enhance robustness and long-horizon performance.
On-policy methods (e.g., PPO~\cite{PPO}) update from freshly collected rollouts (e.g., LUMOS~\cite{lumos}, VLA-RL~\cite{vla_rl}, RoboCLIP~\cite{roboclip},World-Env~\cite{world_env}, RobustVLA~\cite{robust_vla}, EUREKA~\cite{eureka}, Refined Policy Distillation~\cite{refined_policy_distill}); 
off-policy methods (e.g., SAC~\cite{SAC}) exploit replay for sample efficiency (e.g., ConRFT~\cite{conrft}, SERL~\cite{serl}, RL-VLM-F~\cite{rl_vlm_f}), with HIL-SERL~\cite{hil_serl} further leveraging human demos and online corrections.
Beyond these on- and off-policy paradigms, the latest $\pi^{*}_{0.6}$~\cite{pistar06} introduces RECAP (RL with Experience and Corrections via Advantage-Conditioned Policies), a scalable RL framework for large VLA models that incorporates advantage-conditioned policy extraction into flow-matching/diffusion-based VLAs, enabling stable and scalable training without relying on complex RL objectives such as PPO.

\subsubsection{Dateset}
\label{sec:2.6}
Recent progress in embodied intelligence is driven by a shift toward data-centric development, where the scale, diversity, and quality of training data largely determine a VLA model’s generalization and robustness.  VLA datasets form a diverse and evolving ecosystem, each providing complementary supervision signals for different aspects of embodied reasoning and control. Tab.~\ref{tab:vla_datasets_double} is an overview of representative embodied datasets. This section categorizes major datasets by their core properties and primary research roles.

\noindent\textbf{(1) Simulation-Centered Datasets.}
These datasets are collected in controlled virtual environments, which support large-scale, safe, and reproducible data generation with full access to state information. This makes them well suited for studying high-level reasoning and long-horizon planning. ALFRED~\cite{alfred} provides expert demonstrations for 25,000 language-grounded household tasks in AI2-THOR and emphasizes long-horizon compositionality. LIBERO~\cite{libero} targets lifelong robot learning by offering procedurally varied tasks that evaluate incremental skill acquisition and retention. Recent datasets such as VLA-3D~\cite{vla_3d} incorporate detailed 3D scene representations paired with language instructions to support the development of 3D-aware vision–language–action models.

\noindent\textbf{(2) Real-World Robotic Manipulation Datasets.}
These datasets are collected from real robotic systems and capture the full complexity of real-world sensing, dynamics, and environmental variability. They are essential for training policies that remain robust under uncertainty and can generalize to unstructured settings. BridgeData V2~\cite{bridgedata_v2} provides large-scale multi-task demonstrations collected across institutions using a standardized single-arm platform and serves as a central resource for generalist manipulation learning. 
DROID~\cite{droid} expands task and environment diversity by offering more than 350K in-the-wild trajectories gathered from 50+ real environments with a low-cost mobile manipulator. 
AgiBot World~\cite{agitbot_colosseo} further increases scale with a million-level corpus spanning broad task and object variations to support large VLA model training. 
Open X-Embodiment (OXE)~\cite{oxe} aggregates over 60 datasets across 22 robot embodiments and currently represents the most comprehensive resource for studying cross-morphology transfer and the emergence of generalist policies.

\noindent\textbf{(3) Human-Centric and Egocentric Datasets.}
These datasets capture data from a first-person human perspective. Although they typically do not include robotic action labels, they are crucial for grounding perception in human experience and for learning to infer human intent. 
The primary approach involves large-scale, egocentric video collection. 
Ego4D~\cite{ego4d} provides thousands of hours of egocentric video that support pretraining visual representations for human-object interaction, which can be effectively transferred to robotic policies. More specialized datasets further enhance this capability. HD-EPIC~\cite{hd_epic} offers detailed, annotated egocentric recordings of unscripted kitchen activities, and HOI4D~\cite{hoi4d} captures 4D human interactions with diverse objects, enabling fine-grained modeling of interaction dynamics. TEACH~\cite{teach} shifts the focus to instruction following by collecting dialogue-driven task execution data, making it a valuable resource for training agents that can collaborate with humans and resolve ambiguities through communication.

\noindent\textbf{(4) Embodied Visual Question Answering Datasets.}
Embodied VQA datasets pair visual scenes with language-based question-answer supervision and are increasingly used to train VLA models that require semantic alignment and environment-level reasoning. 
MT-EQA~\cite{mutli_eqa} provides 19,287 QA pairs for multi-target embodied question answering, requiring agents to navigate 3D indoor environments and infer object attributes through active exploration. 
EgoTaskQA~\cite{egotaskqa} expands cognitive scope with 368K generated questions refined into 40K high-quality pairs covering description, prediction, explanation, and counterfactual reasoning. 
EmbodiedEval~\cite{embodied_eval} further broadens task diversity with 328 embodied tasks across 125 scenes, including Attribute QA on object and scene properties and Spatial QA that evaluates spatial reasoning through interaction and observation.

\subsubsection{Evaluation} 
\label{sec:2.7}
Standardized benchmarks are central to embodied intelligence research because they define common evaluation protocols that support fair comparison, systematic diagnosis of model limitations, and reproducible experimentation.  The current VLA benchmark ecosystem is diverse, with platforms tailored to assess different dimensions of embodied competence, from basic skills to advanced cognitive abilities.  
Tab.~\ref{tab:vla_benchmarks} is an overview of representative embodied benchmarks.
This section reviews major benchmarks and categorizes them by the primary capabilities they aim to evaluate.

\noindent\textbf{(1) Language-Conditioned Manipulation.}
This category evaluates a model’s ability to follow natural language instructions and produce precise manipulation actions. RLBench~\cite{rlbench} provides over 100 language-annotated tasks with motion-planned demonstrations and serves as a standard benchmark for imitation and reinforcement learning. The ManiSkill series (ManiSkill~\cite{maniskill}, ManiSkill2~\cite{maniskill2}, ManiSkill-HAB~\cite{maniskill_hab}) offers large-scale simulation environments designed to assess multi-task manipulation and policy generalization, with ManiSkill-HAB providing high-fidelity home-environment tasks. RoboMimic~\cite{robominic} evaluates offline learning methods using human demonstrations and highlights key challenges in leveraging human-generated data for manipulation policies.

\noindent\textbf{(2) Long-Horizon and Interactive Task Completion.}
This category evaluates tasks that require sequential reasoning, memory, and sustained interaction with the environment or a human user. 
ALFRED~\cite{alfred} assesses long-horizon compositional household tasks involving irreversible state changes, which challenge planning, memory, and instruction following. CALVIN~\cite{calvin} links language commands with continuous control and evaluates an agent’s ability to execute long sequences of language-guided operations in unseen environments while maintaining state and performing sequential reasoning.
TEACH~\cite{teach} advances toward interactive task execution by introducing dialogue-based instruction following, where the agent must seek clarification and recover from errors through natural language communication.

\noindent\textbf{(3) Advanced Cognitive Capabilities.}
This category includes benchmarks designed to probe higher-level cognitive functions beyond basic instruction following, such as lifelong learning and physical reasoning.
LIBERO~\cite{libero} quantifies lifelong learning dynamics through forward and backward transfer metrics that measure how an agent acquires new skills and retains prior ones across a task sequence. RoboCAS~\cite{robocas} probes embodied cognition in cluttered and physically unstable scenes, exposing the limitations of current models in physical reasoning, spatial understanding, and robust interaction with unpredictable environments.

\noindent\textbf{(4) Evaluation of Embodied Foundation Models.}
This category shifts the evaluation focus from single-task agents to the holistic and emergent capabilities of large pretrained multimodal systems. EmbodiedBench~\cite{embodiedbench} evaluates multimodal large language models such as GPT-4o across high-level semantic planning and low-level physical control to diagnose their end-to-end embodied competence. 
EWMBench~\cite{ewmbench} measures the physical realism of generative world models by assessing the motion and semantic consistency of their predicted futures. RoboTwin~\cite{robotwin} targets cross-robot generalization and evaluates policies on dual-arm collaborative tasks, emphasizing their ability to transfer from large-scale synthetic data.

\input{tables/dataset_table}
\input{tables/benchmark_table}
\input{tables/timeline_table}

%% file: tables/dataset_table.tex
% ==== Slim TPAMI-style: outer border only + vertical centering ====
% 需要 array 宏包：\usepackage{array}
\newcolumntype{L}[1]{>{\raggedright\arraybackslash}m{#1}}   % 左对齐 + 垂直居中
\newcolumntype{C}[1]{>{\centering\arraybackslash}m{#1}}     % 居中 + 垂直居中

\begin{table*}[!t]
\begin{center}
\caption{\textbf{An overview of representative embodied datasets}. We exhibit different facets of these datasets, including embodiment, perspective, episodes, scenes, tasks\&skills, and collection. More details are discussed in Section~\ref{sec:2.6}.}
\label{tab:vla_datasets_double}
\scriptsize
\setlength{\tabcolsep}{3pt}
\renewcommand{\arraystretch}{1.12}

\begin{tabular}{
 L{2.40cm}  % Name (Year)
  L{2.55cm}  % Embodiment
  L{2.30cm}  % Perspective
  L{2.55cm}  % Episodes
  L{2.10cm}  % Scenes
  L{2.50cm}  % Tasks & Skills
  L{2.55cm}  % Collection
}
\hline
\textbf{Name (Year)} &
\textbf{Embodiment} &
\textbf{Perspective} &
\textbf{Episodes} &
\textbf{Scenes} &
\textbf{Tasks \& Skills} &
\textbf{Collection} \\

\midrule
\multicolumn{7}{c}{\textbf{Simulation-Centered Datasets}} \\
\midrule

ALFRED~\cite{alfred} (2020) & Simulated human agent & First-person &
8,055 expert demonstrations & $\sim$120 indoor scenes &
8 composite household activities & Simulation (AI2-THOR) \\
LIBERO~\cite{libero} (2022) & Simulated robot arm & First-person &
$\sim$6,500 & 4 simulated domains & 130 skills & Simulation (Robosuite) \\
VLA-3D~\cite{vla_3d} (2024) & Virtual agent in 3D scenes & Third-person &
9.7M referential pairs & 11.5k reconstructed 3D rooms &
Spatial navigation \& grounding & Simulation (Matterport3D / ScanNet) \\

\midrule
\multicolumn{7}{c}{\textbf{Real-World Robotic Manipulation Datasets}} \\
\midrule
BridgeData V2~\cite{bridgedata_v2} (2023) & Robot arm (WidowX) & Mixed (first- \& third-person) &
60,096 trajectories & 24 real environments &
13 core manipulation skills & Real robot (VR teleoperation + scripted) \\
DROID~\cite{droid} (2024) & Robot arm (Franka Emika Panda) &
Mixed (wrist \& external cameras) &
$\sim$76k ($\approx$350 hours) & 564 distinct real scenes &
86 tasks & Real robot (VR teleoperation by 50 operators) \\
Open X-Embodiment~\cite{oxe} (2023) & 22 robot types  &
Mixed (first- \& third-person) & 1M+ trajectories &
160k+ unified scenes & 527 skills &
Web-scale aggregation of real-robot data \\
AgiBot World~\cite{agitbot_colosseo} (2024) & Dual-arm humanoid robot fleet & First-person &
1M+ trajectories & 5 domains (home, retail, office, restaurant, industry) &
217 tasks & Real robot (multi-robot facility) \\

\midrule
\multicolumn{7}{c}{\textbf{Human-Centric and Egocentric Datasets }} \\
\midrule
Ego4D~\cite{ego4d} (2021) & Human & First-person &
$\sim$3{,}700 hours ($\sim$1M clips) & 74 locations across 9 countries &
Multi-activity & Real human egocentric video \\
TEACh~\cite{teach} (2021) & Human commander + embodied agent &
Mixed (first- \& third-person) &
$\sim$3k dialog-based episodes & $\sim$200 simulated homes &
17 composite household tasks & Human teleoperation in simulation \\
HOI4D~\cite{hoi4d} (2022) & Human & First-person &
$\sim$4,000 sequences & 610 indoor scenes &
54 tasks across all 16 categories & Head-mounted dual RGB-D \\
HD-EPIC~\cite{hd_epic} (2025) & Human & First-person &
$\sim$4{,}881 object itineraries & 9 Real kitchen scenes &
-- & Wearable sensors (Project Aria glasses) \\

\midrule
\multicolumn{7}{c}{\textbf{Embodied Visual Question Answering Datasets}} \\
\midrule
MT-EQA~\cite{mutli_eqa} (2019) & -- & First-person &
$\sim$19{,}287 QA pairs & 588 environments &
61 unique object in 8 unique room  & Simulation (House3D) \\
EgoTaskQA~\cite{egotaskqa} (2022) & Human & First-person &
$\sim$40K QA pairs & Kitchen &
48 relationships and 14 object attributes  & Head-mounted egocentric RGB video \\
EmbodiedEval~\cite{embodied_eval} (2025) & -- & First-person &
328 tasks & 125 unique scenes &
Navigation Spatial,Attribute, \ldots  & -- \\
\hline
\end{tabular}
\end{center}
\end{table*}

%% file: tables/benchmark_table.tex
% 导言区
\renewcommand\arraystretch{1.12}
\setlength{\tabcolsep}{3pt}

\begin{table*}[!t]
\centering
\scriptsize
\setlength{\tabcolsep}{3pt}
\renewcommand{\arraystretch}{1.12}
\caption{\textbf{An overview of representative embodied benchmarks}. We exhibit different facets of these benchmarks, including task type, evaluation metrics, and environment/platform. More details are discussed in Section~\ref{sec:2.7}.}
\begin{tabularx}{\textwidth}{
    >{\raggedright\arraybackslash}p{3.0cm}  % Name (Year)
    >{\raggedright\arraybackslash}p{3.5cm}  % Task Type
    X  % Evaluation Metric
    X  % Environment / Platform
}
\toprule
\textbf{Name (Year)} &
\textbf{Task Type} &
\textbf{Evaluation Metric} &
\textbf{Environment / Platform} \\
\midrule
\addlinespace[3pt]
\multicolumn{4}{c}{\textbf{Language-Conditioned Manipulation \& Control}} \\
\midrule
\addlinespace[2pt]
RLBench~\cite{rlbench} (2020) &
Multi-task tabletop manipulation &
Success rate &
PyRep / CoppeliaSim \\

\addlinespace[1pt]
ManiSkill Series~\cite{maniskill,maniskill2,maniskill_hab} &
Multi-task object-centric manipulation &
Success / completion rate (per task) &
SAPIEN (ManiSkill), Habitat-based (ManiSkill-HAB) \\

\addlinespace[1pt]
RoboMimic~\cite{robominic} (2021) &
Multi-stage robot manipulation &
Success rate &
MuJoCo \\
\midrule

\addlinespace[3pt]
\multicolumn{4}{c}{\textbf{Long-Horizon and Interactive Task Completion}} \\
\midrule

\addlinespace[2pt]
ALFRED~\cite{alfred} (2020) &
Vision–language instruction following &
Success rate, Goal-Condition Success &
ALFRED simulator \\

\addlinespace[1pt]
CALVIN~\cite{calvin} (2022) &
Language-guided multi-step manipulation &
Success rate, zero-shot generalization &
Simulated tabletop (4 scenes) \\

\addlinespace[1pt]
TEACh~\cite{teach} (2021) &
Dialog-driven embodied task completion &
Success rate, EDH / TfD / TATC &
AI2-THOR \\

\midrule
\addlinespace[3pt]
\multicolumn{4}{c}{\textbf{Advanced Cognitive Capabilities}} \\
\midrule

\addlinespace[2pt]
LIBERO~\cite{libero} (2023) &
Continual multi-task manipulation &
Success rate, Fwd/Bwd Transfer, AUC &
Robosuite \\

\addlinespace[1pt]
RoboCAS~\cite{robocas} (2024) &
Multi-object arrangement \& long-horizon manipulation &
Success under spatial/clearance constraints &
Custom arrangement scenes (SAPIEN) \\

\midrule
\addlinespace[3pt]
\multicolumn{4}{c}{\textbf{Evaluation of Embodied Foundation Models}}\\
\midrule

\addlinespace[2pt]
EmbodiedBench~\cite{embodiedbench} (2025) &
Vision-driven embodied agent evaluation &
Success rate, Subgoal success rate &
AI2-THOR, Habitat 2.0, CoppeliaSim \\

\addlinespace[1pt]
EWM Bench~\cite{ewmbench} (2025) &
World-model evaluation &
Scene consistency, motion correctness, semantic alignment &
Synthetic + real embodied datasets \\

\addlinespace[1pt]
RoboTwin~\cite{robotwin} (2025) &
Multi-robot imitation, cross-embodiment manipulation &
Success rate, sim\,$\leftrightarrow$\,real transfer rate, latency &
Isaac Gym / PyBullet \\
\bottomrule
\end{tabularx}
\label{tab:vla_benchmarks}
\end{table*}

%% file: tables/timeline_table.tex
% 需要 \usepackage{graphicx}（IEEEtran 默认已载）
\begin{table*}[!htbp]
\begin{center}
\caption{\textbf{An overview of VLA milestones}. We exhibit different facets of these methods, including robot perception, brain, action, training strategy, primary dataset, and evaluation, which corresponding to the subsections in Section~\ref{sec:3}.}
\label{tab:vla_timeline_big}
\scriptsize
\setlength{\tabcolsep}{1.6pt}   % 稍缩内边距，避免拥挤
\renewcommand\arraystretch{1.12}

\resizebox{\textwidth}{!}{%
\begin{tabular}{p{2.25cm} p{3.5cm} p{3.00cm} p{3.0cm} p{1.60cm} p{2.2cm} p{1.5cm}}
\hline
\textbf{Name} & \textbf{\mbox{Perception (Visual/Language)}} & \textbf{Brain} &
\textbf{Action} & \textbf{Training} & \textbf{\mbox{Primary Dataset}} & \textbf{Evaluation} \\
\hline

\multicolumn{7}{l}{\textbf{By 2021}} \\
EmbodiedQA~\cite{embodiedqa} & CNN/LSTM & LSTM+FNN& Discrete(Autoregressive) & BC & EQA dataset & EQA v1 \\
VLN~\cite{vln} & ResNet-152 / LSTM & LSTM & Discrete(Autoregressive) & BC & R2R & R2R \\
RCM~\cite{rcm} & ResNet-152/LSTM & LSTM  & Discrete(Autoregressive) & BC + RL & R2R & R2R \\
Point-Cloud EQA~\cite{point_cloud_eqa} & PointNet++ \& ResNet50/LSTM & RRN+GRU-RNN & Discrete (Sequentia) & BC & MP3D-EQA & Matterport3D \\
ALFWorld~\cite{alfredworld} & Mask R-CNN / Seq2Seq & Seq2Seq & Discrete(Autoregressive) & BC & TextWorld & ALFRED benchmark \\
CLIPort~\cite{cliport} &  ResNet-50 / Transformer  & FCN + Affordances & Discrete(Autoregressive) & BC & Ravens  & Ravens  \\
\hline

\multicolumn{7}{l}{\textbf{2022}} \\
SayCan~\cite{saycan} & Resnet-18 / LLM & LLM & Discrete(Autoregressive) & BC + RL & -- & -- \\
Inner\\Monologue~\cite{inner_monologue} & LLM & LLM & Discrete(Autoregressive) & BC & Everyday Robots, Ravens   \& Ravens & Ravens  \\
RT-1~\cite{rt_1} & EfficientNet-B3 / USE & Transformer & Discrete(Autoregressive) & BC & Self-collected & Self-built benchmark \\
RT-2~\cite{rt_2} & PaLI-X + PaLM-E & VLM & Discrete(Autoregressive) & BC + co-finetuning & WebLI + RT-1 & Real-world Generalization Benchmark \\
\hline

\multicolumn{7}{l}{\textbf{2023}} \\
PaLM-E~\cite{palm_e} & ViT / PaLM  & VLM & Discrete(Autoregressive) & Multimodal SFT & WebLI,$\ldots$ & OK-VQA,$\ldots$ \\
Diffusion \\Policy~\cite{diffusion_policy} & ResNet-18 & Transformer/DiT & Continuous(DDPM) & BC & Human demonstration data & Robomimic,Push-T,$\ldots$ \\
\hline

\multicolumn{7}{l}{\textbf{2024}} \\
3D-VLA~\cite{3d_vla} & VLM &  3D-LLM  & Continuous(trajectory segment) & co-finetuning & OXE, RLBench, $\ldots$ & RLBench, RoboVQA,$\ldots$ \\
Octo~\cite{octo} & CNN / T5 & Transformer & Continuous(DDPM) & BC & OXE & Policy generalization, SR \\
OpenVLA~\cite{openvla} & SigLip + Dino & Transformer & Discrete(Autoregressive) & BC & OXE & Libero \\
GR-2~\cite{gr_2} & VQGAN / CLIP & Transformer & Continuous & Predictive modeling & HowTo100M, Ego4D,$\ldots$ & CALVIN, $\ldots$ \\
$\pi_0$~\cite{pi0} & VLM & Transformer  & Continuous(Flow Matching) & BC & OXE, Bridge v2,$\ldots$ & -- \\
\hline

\multicolumn{7}{l}{\textbf{2025}} \\
Humanoid-VLA~\cite{humanoid_vla} & Transformer & Transformer & Continuous(Autoregressive) & BC & Humanoid-S, AMASS & HumanML3D, Humanoid-S,$\ldots$ \\
GR00T N1~\cite{groot_n1} & Eagle-2 VLM & VLM + DiT & Continuous(Flow Matching) & BC & GR00T N1 dataset, OXE,$\ldots$ & GR-1 Tabletop Tasks,$\ldots$ \\
PointVLA~\cite{pointvla} & CNN (3D) / VLM & VLM & Continuous(DDPM) & BC & Self-collected data  & RoboTwin \\
CoT-VLA~\cite{cot_vla} & Transformer & LLM & Discrete & BC & OXE & Libero, Brige v2, $\ldots$  \\
$\pi_{0.5}$~\cite{pi0_5} & VLM & Transformer & Hybrid(Flow Matching) & BC + Predictive modeling & OXE & Real family scenes \\
LUMOS~\cite{lumos} & CNN / Sentence-BERT & Goal-conditioned actor-critic & Continuous/ & BC & -- & MLPerf Training Benchmarks \\
VLA-RL~\cite{vla_rl} & Siglip + Dino / Llama-2  & Transformer & Discrete(Autoregressive) & RL & Online-collected & LIBERO  \\ 
Cosmos-R1~\cite{cosmos_r1} & ViT/LLM & LLM & Discrete(Autoregressive) & BC + RL & BridgeData v2,RoboVQA,... & RoboFail,AgiBot,... \\
\hline
\end{tabular}
}% end resizebox
\end{center}
\end{table*}